\documentclass[10pt,twocolumn,letterpaper]{article}

\usepackage[pagenumbers]{cvpr} 

\definecolor{cvprblue}{rgb}{0.21,0.49,0.74}
\usepackage[pagebackref,breaklinks,colorlinks,allcolors=cvprblue]{hyperref}
\usepackage{graphicx, stfloats}
\usepackage{animate}
\usepackage{cuted}
\usepackage{tabularx}
\usepackage{capt-of}

\usepackage{multirow}
\usepackage{multicol}
\usepackage{titletoc}

\usepackage[linesnumbered,ruled,vlined]{algorithm2e}

\usepackage[most]{tcolorbox} 
\usepackage{xcolor} 
\usepackage{lipsum} 

\tcbset{
    mybox/.style={
        colback=white,          
        colframe=gray!60,       
        fonttitle=\bfseries,    
        coltitle=black,         
        rounded corners,        
        boxrule=1pt,            
        enhanced,
        width=\textwidth,       
        arc=6mm,                
        frame style={opacity=0.8}, 
        breakable               
    }
}

\def\paperID{6015} 
\def\confName{CVPR}
\def\confYear{2025}

\title{Spatiotemporal Skip Guidance for Enhanced Video Diffusion Sampling}

\newcommand\blfootnote[1]{%
  \begingroup
  \renewcommand\thefootnote{}\footnote{#1}%
  \addtocounter{footnote}{-1}%
  \endgroup
}
\newcommand*{\affaddr}[1]{#1}
\newcommand*{\affmark}[1][*]{\textsuperscript{#1}}
\newcommand*{\email}[1]{\texttt{#1}}
\author{
Junha Hyung$^{*}$\affmark[1]\quad Kinam Kim$^{*}$\affmark[1]\quad Susung Hong\affmark[2]\quad Min-Jung Kim\affmark[1]\quad Jaegul Choo\affmark[1]\\
\\
\affaddr{\affmark[1]KAIST AI}\quad \affaddr{\affmark[2]University of Washington}\\
\small\email{\{sharpeeee, kinamplify, emjay73, jchoo\}@kaist.ac.kr, susung@cs.washington.edu}
\vspace{-1cm}
}

\begin{document}
\maketitle

\begin{strip}\centering
    \newcommand{\numColumns}{2}
    \newcommand{\columnSpacing}{0.em}
    \begin{tabular}{
        @{}
        p{\dimexpr(\textwidth-\columnSpacing*(\numColumns-1))/\numColumns} @{\hspace{\columnSpacing}}
        p{\dimexpr(\textwidth-\columnSpacing*(\numColumns-1))/\numColumns} @{}
    }
        \animategraphics[loop, width=\linewidth]{8}{videos/combined_5/}{0000}{90} &
        \animategraphics[loop, width=\linewidth]{8}{videos/combined_10/}{0000}{90}
    \end{tabular}
    \begin{tabularx}{\textwidth}{XX}
        \centering {\footnotesize\fontfamily{put}\selectfont \textit{"A close-up shot of a butterfly landing on the nose of a woman, highlighting her smile and the details of the butterfly's wings."}} & 
        \centering {\footnotesize\fontfamily{put}\selectfont \textit{"A close-up of a woman's face with colored powder exploding around her, creating an abstract splash of vibrant hues."}}
    \end{tabularx}
    \captionof{figure}{Visual comparison of video quality between CFG (top row) and our STG method (bottom row). Best viewed in Acrobat Reader; click on the images to watch the videos.}
    \label{fig:main}
\end{strip}

\blfootnote{* indicates equal contribution.}
\label{sec:Abstract}
\begin{abstract}

Diffusion models have emerged as a powerful tool for generating high-quality images, videos, and 3D content. While sampling guidance techniques like CFG improve quality, they reduce diversity and motion. Autoguidance mitigates these issues but demands extra weak model training, limiting its practicality for large-scale models.
In this work, we introduce Spatiotemporal Skip Guidance (STG), a simple training-free sampling guidance method for enhancing transformer-based video diffusion models.
STG employs an implicit weak model via self-perturbation, avoiding the need for external models or additional training. 
By selectively skipping spatiotemporal layers, STG produces an aligned, degraded version of the original model to boost sample quality without compromising diversity or dynamic degree. 
Our contributions include: (1) introducing STG as an efficient, high-performing guidance technique for video diffusion models, (2) eliminating the need for auxiliary models by simulating a weak model through layer skipping, and (3) ensuring quality-enhanced guidance without compromising sample diversity or dynamics unlike CFG. For additional results, visit \url{https://junhahyung.github.io/STGuidance/}.

\end{abstract}

\section{Introduction}
\label{sec:intro}

Diffusion models~\cite{sohl2015deep, ho2020denoising, song2020score, rombach2022high} are a successful class of generative models known for their flexibility in modeling complex data distributions, achieving impressive results in image, video, and 3D generation.
By progressively denoising random noise, they enable robust generalization, making them a leading choice for realistic content generation and often surpassing GAN-based methods~\cite{goodfellow2020generative, brock2018large, razavi2019generating, karras2020analyzing}.
Building on this success, video diffusion models~\cite{ho2022video, blattmann2023stable, genmo2024mochi, polyak2024movie, opensora} generate high-quality videos by using temporal or 3D attention layers to handle sequential frames.

Meanwhile, to enhance sample quality, sampling guidance techniques such as Classifier-Free Guidance (CFG)~\cite{ho2022classifier} and Autoguidance~\cite{karras2024guiding} have been introduced to guide the denoising process.
These techniques employ \textit{weak models} to predict poor trajectories, steering the main model away from them and pushing samples toward high-quality regions on the data manifold. 
However, CFG often reduces diversity, leading to saturated or overly simplified results~\cite{karras2024guiding, chung2024cfg++}.

Autoguidance~\cite{karras2024guiding} addresses this issue by using a weak model trained on the same task, conditioning, and data distribution as the main model. 
The main drawback of this approach, however, is the need to train an additional weak model, which is impractical for large-scale models. 
Alternative methods such as PAG~\cite{ahn2024self} and SEG~\cite{hong2024smoothed} use self-perturbation to implicitly mimic a weak model, avoiding the need for extra training. 
Yet, these methods are designed specifically for image generation diffusion models, applying self-perturbation to 2D spatial attention maps.

In this work, we propose Spatiotemporal Skip Guidance (STG), a simple and effective sampling guidance method for video diffusion models that significantly enhances the performance of any transformer-based video diffusion model without additional training. Specifically, we use an implicit weak model for guidance through self-perturbation, eliminating the need for explicit weak models and their associated training costs. This is especially crucial for video diffusion models, where training costs are high.

\begin{figure}
    \centering
    \includegraphics[width=\linewidth]{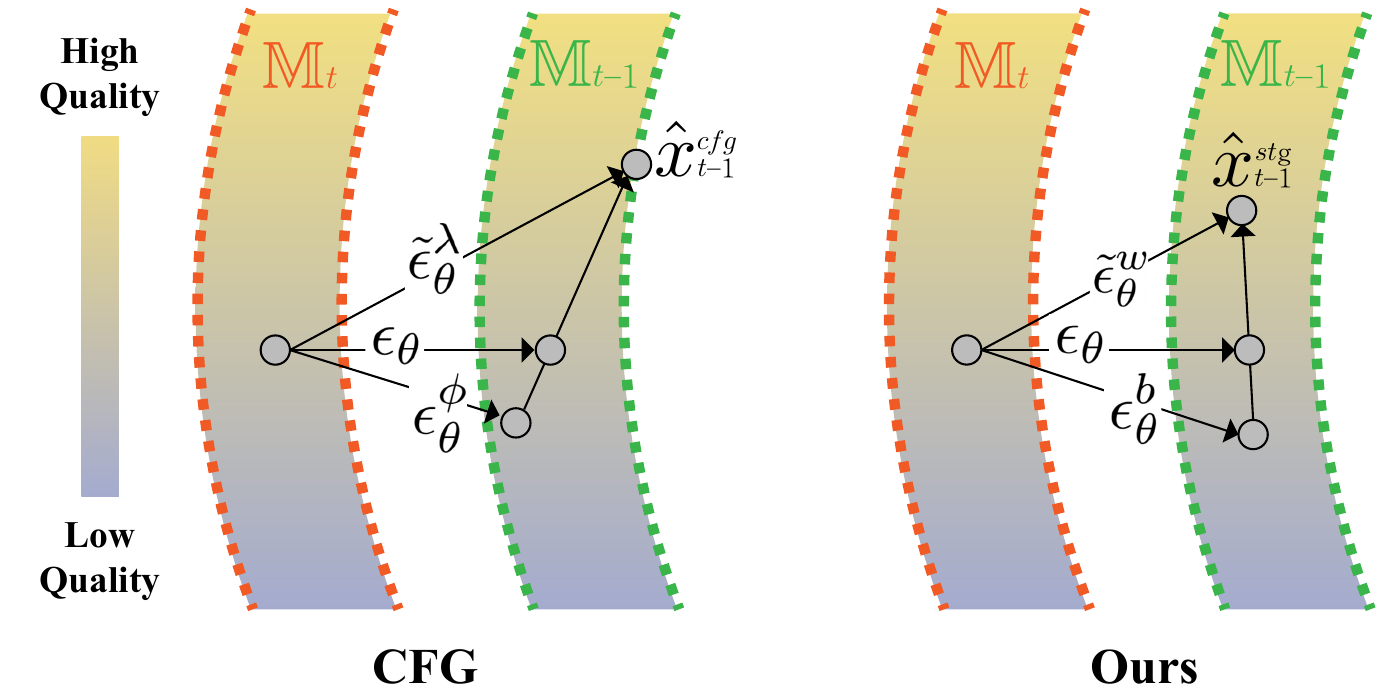}
    \caption{Comparison between CFG and STG, with the band conceptually representing the noisy data manifold. In STG, the weak model and the main model are aligned along the direction of increasing quality. In contrast, the two models in CFG differ not only in quality but also in aspects such as diversity and prompt alignment capabilities.}
    \label{fig:conceptual}
    \vspace{-6mm}
\end{figure}

Our implicit weak model, a key component of our framework, is deliberately designed as a degraded but ``aligned'' version of the original video generation model. As demonstrated in Autoguidance~\cite{karras2024guiding}, having the weak model share the same task, conditioning, and data distribution as the main model is essential for quality improvement, as we expect both models to exhibit similar, aligned errors in the same parts of samples. We conceptually illustrate this alignment in Fig.~\ref{fig:conceptual}, where our weak model and main model are arranged in a direction of increasing quality. To achieve this, we apply spatiotemporal perturbation to both spatial and temporal attention layers---or, in the case of 3D attention, to the entire layer---by selectively \textit{skipping} certain layers. This straightforward approach effectively nullifies specific residual or attention layers, generating a lower-quality version of the model that simulates an aligned weak model.


To ensure samples remain on the data manifold, even when using large guidance scales, we employ optional techniques such as the rescaling~\cite{lin2024common} and restart~\cite{xu2023restart} methods. Rescaling the latent code constrains its variance, addressing the issue of larger variance causing saturation in the results~\cite{lin2024common}. Meanwhile, the restart sampling technique leverages the error contraction property of forward SDEs~\cite{xu2023restart} to keep the sampling trajectory on the manifold. These techniques help prevent overshooting in the sampling trajectory, which could otherwise push samples off the manifold and result in saturated or distorted outputs.

Our key contributions are as follows: 
\begin{itemize} \item We propose STG—a surprisingly simple sampling guidance framework that significantly boosts the performance of video diffusion models. 
\item Our method introduces an implicit weak model by skipping spatiotemporal layers in video diffusion models, eliminating the need for additional training or external models. 
\item Our method enhances sample quality during guidance without reducing diversity or limiting the dynamics of generated videos.
\end{itemize}

\section{Related Work}
\label{sec:related_work}



\paragraph{Guidance with trained weak model}
Classifier-Free Guidance (CFG)~\cite{ho2022classifier} improves conditional generation in Diffusion Models by using an implicit unconditional model as a weak model. However, differences in tasks between the unconditional and conditional models can reduce sample diversity~\cite{karras2024guiding, chung2024cfg++} and increase sampling trajectory curvature~\cite{chung2024cfg++}, leading to overshooting the data manifold and producing skewed or oversaturated images.

Autoguidance~\cite{karras2024guiding} mitigates these issues by employing a \textit{bad version of the main model} as a weak model, trained with reduced capacity and compute to ensure alignment. This alignment allows the guidance algorithm to correct errors by analyzing prediction differences. While effective, it requires additional training, which is challenging for large-scale video diffusion models.


\paragraph{Guidance with training-free weak model}
Another line of work avoids additional training by using self-perturbation of the main model to mimic a weak model.
Self-Attention Guidance (SAG)~\cite{hong2023improving} blurs high-attention regions, Perturbed Attention Guidance (PAG)~\cite{ahn2024self} replaces attention maps with identity matrices, and Smoothed Energy Guidance (SEG)~\cite{hong2024smoothed} applies Gaussian blur to the attention weights to smooth the energy landscape. These methods guide sampling toward high-quality outputs by leveraging differences in predictions from the weakened model. While effective, they are primarily designed for image diffusion models with 2D self-attention. We aim to extend this approach to video diffusion models, which require handling temporal dynamics with additional temporal or 3D spatiotemporal attention layers.

\section{Preliminaries}
\label{sec:Preliminaries}
\subsection{Diffusion models}
Diffusion models generate samples by progressively removing noise from noisy data, restoring the original data distribution.
Song et al.~\cite{song2020score} defined the process of adding noise to the data using a stochastic differential equation (SDE):
\begin{align}
    \label{eq:forward-sde}
    dx = -\frac{\beta(t)}{2}x \, dt + \sqrt{\beta(t)} \, dw,
\end{align}
where \(\beta(t)\) is a time-dependent noise schedule, and \(w\) represents the standard Wiener process.
Corresponding reverse-time SDE is:
\begin{align}
    \label{eq:reverse-sde}
    dx = \left[ -\frac{\beta(t)}{2}x - \beta(t) \nabla_x \log p_t(x) \right] dt + \sqrt{\beta(t)} \, d\bar{w}
\end{align}
where \(d\bar{w}\) is the Wiener process running backward in time. 
Here, the score function \(\nabla_x \log p_t(x)\) is approximated by a neural network $s_\theta(x(t))$ trained using denoising score matching~\cite{vincent2011connection}: 
\begin{multline}
    \label{eq:score-matching}
    \theta^* = \arg \min_\theta \mathbb{E}_t \left\{ \lambda(t) \mathbb{E}_{x_0} \mathbb{E}_{x_t|x_0} \left[  \right. \right. \\
    \left. \left. \|s_\theta(x_t)- \nabla_{x_t} \log p_t(x_t | x_0) \|_2^2 \right] \right\}.
\end{multline}

\subsection{Classifier Guidance}
Classifier Guidance (CG)~\cite{mukhopadhyay2023diffusion} reformulates the reverse process of a diffusion model by incorporating an external classifier \( p_\phi \) as
\begin{align}
    \label{eq:cg-distribution}
    \tilde{p_\theta}(x_t | y) \propto p_\theta(x_t) p_\phi(y | x_t)^{\lambda},
\end{align}
where \( y \) is the desired class label and $\lambda$ controls the guidance strength. 
The score function is then derived as:
\begin{multline}
    \label{eq:cg-scorefunction}
    \nabla_{x_t} \log \tilde{p}_\theta(x_t | y) = \nabla_{x_t} \log p_\theta(x_t) \\ + \lambda\nabla_{x_t}\log p_\phi(y | x_t).
\end{multline}
By substituting the score function in Eq.~\ref{eq:reverse-sde} with Eq.~\ref{eq:cg-scorefunction}, sampling from the desired class condition $y$ becomes possible.

\subsection{Classifier-Free Guidance}
Classifier-Free Guidance (CFG)~\cite{ho2022classifier} uses Bayes' rule to replace a classifier-guided score with a linear combination of conditional and unconditional score estimates:
\begin{align}
    \Tilde{\epsilon}^{\lambda}_\theta(x_t) = \epsilon_\theta(x_t) + \lambda \left( \epsilon_\theta(x_t) - \epsilon_\theta(x_t | \phi) \right).
\end{align}
CFG jointly trains the unconditional model $\epsilon_\theta(x_t | \phi)$ and the conditional model $\epsilon_\theta(x_t|c)$ $(= \epsilon_\theta(x_t))$ within a single model by setting the condition $c$ to a null token $\phi$. 
Using this guided denoising process, the model better captures the conditions and often generates high-fidelity outputs.


\section{Method}
\label{sec:method}

%
\subsection{Optimal Weak Model Design}
\label{sec:optimal_weak_model_design}

Our goal is to construct an \textit{aligned} weak model that captures a distribution similar to the original model while generating slightly lower-quality samples. This ensures that the guidance gradient points toward improved quality, as shown in Fig.~\ref{fig:conceptual}. Misaligned models, as seen in CFG~\cite{ho2022classifier}, may lead to unintended outcomes like reduced diversity, while Autoguidance~\cite{karras2024guiding} requires additional weak model training, making it impractical for large-scale models. 

To address these limitations, we design an \textit{implicit} weak model by leveraging the original model itself, eliminating the need for external training. This training-free approach ensures computational efficiency and better alignment, as the weak model is derived directly from the original, preserving most network weights.

To achieve this, we apply perturbation methods directly to the main model’s forward pass, creating an implicit weak model. Specifically, we use \textit{spatiotemporal} perturbation, a key component for aligning the weak model with video diffusion models, which will be elaborated upon in detail.



\subsection{Sampling from High Quality Samples}


Similar to CG~\cite{mukhopadhyay2023diffusion}, we define our goal as conditioning the model on an imaginary label $y_g$ that represents high-quality samples, leading to the sampling distribution
\begin{equation}
\Tilde{p}_{\theta}(x_t|y_g) \propto {p}_{\theta}(x_t) {{p}_{\phi}(y_g|x_t)}^w.
\label{eq:cg_ours}
\end{equation}
Using $w>0$ sharpens the distribution, promoting the generation of high-quality samples.
From Eq.~\ref{eq:cg_ours}, the score is derived as
\begin{multline} \label{eq:grad_log_bayes}
\nabla_{x_t} \log \Tilde{p}_{\theta}(x_t|y_g) \\ = 
\nabla_{x_t} \log p_{\theta}(x_t) + w\nabla_{x_t} \log p_{\phi}(y_g|x_t).
\end{multline}
Rather than using an external classifier $p_{\phi}$, we propose using our model $p_{\theta}$ as an implicit classifier.
We design this classifier to be inversely proportional to the probability of an imaginary ``bad" label $y_b$, expressed as
\begin{equation} 
p_{\phi}(y_g|x_t) \propto \frac{1}{p_{\theta}(y_b | x_t)}.
\label{eq:deriv1}
\end{equation}

\begin{table*}[t]
\centering
\resizebox{0.95\textwidth}{!}{%
\begin{tabular}{lccccc}
\toprule
\textbf{Models}      & Imaging Quality & Aesthetic Quality & Motion Smoothness & Dynamic Degree & Temporal Flickering \\
\midrule
Mochi (CFG) &  0.524 & 0.507 & 0.985 & \textbf{0.87} & 0.976\\
Mochi (STG) & \textbf{0.628} & \textbf{0.554} & \textbf{0.988} & 0.86 & \textbf{0.978}\\
\midrule
Open-Sora (CFG) &  0.561 & 0.493 & 0.982 & \textbf{0.902}& 0.975 \\
Open-Sora (STG) & \textbf{0.606} & \textbf{0.509} & \textbf{0.987} &0.895 & \textbf{0.976} \\
\bottomrule
\end{tabular}%
}
\caption{Quantitative results for Mochi~\cite{genmo2024mochi} and Open-Sora~\cite{opensora} on VBench~\cite{huang2024vbench} T2V benchmarks.
}
\label{tab:quant-t2v}
\end{table*}

\begin{table*}[t]
\centering
\resizebox{0.8\textwidth}{!}{%
\begin{tabular}{lcccccc}
\toprule
\textbf{Models}    & FVD ($\downarrow$) & IS           & Imaging Quality & Aesthetic Quality & Motion Smoothness & Dynamic Degree  \\
\midrule
SVD (CFG) & 151.3 & 38.0& 0.687 & 0.637 & 0.966 & 0.562  \\

SVD (STG) & \textbf{128.7} & \textbf{38.5}& \textbf{0.694} & \textbf{0.639} & \textbf{0.968} & \textbf{0.694} \\
\bottomrule
\end{tabular}
}
\caption{Quantitative results for SVD~\cite{blattmann2023stable} on FVD, IS, and VBench~\cite{huang2024vbench} I2V benchmarks.}
\label{tab:quant-i2v}
\end{table*}

Using Bayes' rule, the function can be expressed in terms of the marginal posterior as follows:
\begin{align}
&\nabla_{x_t} \log \frac{1}{p_{\theta}(y_b | x_t)}
= \nabla_{x_t} \log \frac{p_{\theta}(x_t)}{p_{\theta}(y_b) p_{\theta}(x_t | y_b)} \notag
\\ &= \nabla_{x_t} (\log p_{\theta}(x_t) - \log p_{\theta}(x_t | y_b)),
\label{eq:deriv2}
\end{align}
leading to the score of the target distribution as:
\begin{multline} \label{eq:score-ours}
\nabla_{x_t} \log \Tilde{p}_{\theta}(x_t|y_g) = 
\nabla_{x_t} \log p_{\theta}(x_t) \\+ w\nabla_{x_t} (\log p_{\theta}(x_t) - \log p_{\theta}(x_t | y_b)).
\end{multline}
We can sample from $\Tilde{p}_{\theta}(x_t|y_g)$ by substituting the score function in Eq.~\ref{eq:reverse-sde} with Eq.~\ref{eq:score-ours}, resulting in:
\begin{align}
\label{eq:reverse-sde-ours}
    dx &= \bigg[-\frac{\beta(t)}{2}x - \beta(t)(\nabla_{x_t} \log p_t(x_t)  \\
     &+ w\nabla_{x_t} (\log p_{\theta}(x_t) - \log p_{\theta}(x_t | y_b))\bigg] dt + \sqrt{\beta(t)} d\bar{w}, \notag
\end{align}
and solving the reverse SDE.
Since the score function is approximated using the neural network, we can generate samples using
\begin{align}
\label{eq:reverse-sde-ours-2}
    \Tilde\epsilon^{w}_{\theta}(x_t) = \epsilon_{\theta}(x_t) + w(\epsilon_{\theta}(x_t) - \epsilon^b_{\theta}(x_t)).
\end{align}

An interesting approach here is to model $\nabla_{x_t} \log p_{\theta}(x_t | y_b)$ by perturbing the forward pass of $\epsilon_{\theta}(x_t)$, denoted as $\epsilon^b_{\theta}(x_t)$.
Now the main focus is designing a perturbation that effectively yields the weak model capable of estimating $\epsilon^b_{\theta}(x_t)$, aligning closely with $\epsilon_{\theta}(x_t)$, as discussed in Sec.~\ref{sec:optimal_weak_model_design}. 
Ideally, we want a distribution that deviates minimally from $\epsilon_{\theta}(x_t)$ while producing slightly lower-quality samples.

\subsection{Spatiotemporal Skip Guidance (STG)}
We introduce Spatiotemporal Skip Guidance (STG), a simple guidance method designed for video diffusion models that generates diverse, high-fidelity samples while addressing the limitations of existing methods. STG implicitly simulates an aligned weak model through spatiotemporal perturbation, capturing the spatiotemporal dynamics of video data.

An interesting discovery of this paper is that \textit{skipping} layers within the network is an effective approach for constructing an aligned weak model. Modern neural network architectures for video diffusion models, such as ADM~\cite{mukhopadhyay2023diffusion}, DiT~\cite{peebles2023scalable}, and SiT~\cite{ma2024sit}, are partially or fully transformer-based and contain attention layers and residual blocks. These architectures are well-suited for layer skipping, as they can still produce plausible outputs without significant degradation, even when a few layers are removed. 

Layer skipping is advantageous for generating an aligned weak model because it retains most of the neural network’s weights and forward pass, resulting in similar predictions and distributions. This approach offers a clear advantage over methods that rely on external models, such as Autoguidance~\cite{karras2024guiding}, or alternative objectives like CFG~\cite{ho2022classifier}, where forward passes differ significantly. 


Moreover, our method applies perturbations to both spatial and temporal layers (or spatiotemporal layers), unlike existing image-based perturbation methods~\cite{hong2024smoothed, hong2023improving} limited to 2D spatial attention layers. This dual-layer perturbation is essential for aligning the weak model with the main video diffusion model. We now discuss various STG configurations that can be applied across different video diffusion model architectures.

\begin{figure*}
    \centering
    {\small \textit{A macro cinematography animation showing a butterfly emerging from its chrysalis, filmed with side-lit lighting ...}}
    \includegraphics[width=\linewidth]{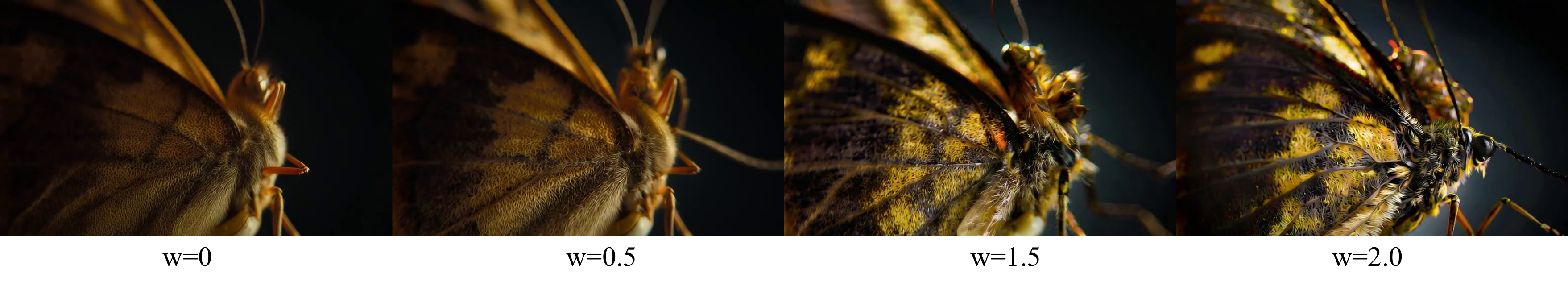}
    \caption{Selected frames from videos generated by Mochi~\cite{genmo2024mochi} with increasing STG scales.
    }
    \label{fig:ablation}
    \vspace{-4mm}
\end{figure*}

\begin{figure}
    \centering
    \includegraphics[width=0.9\linewidth]{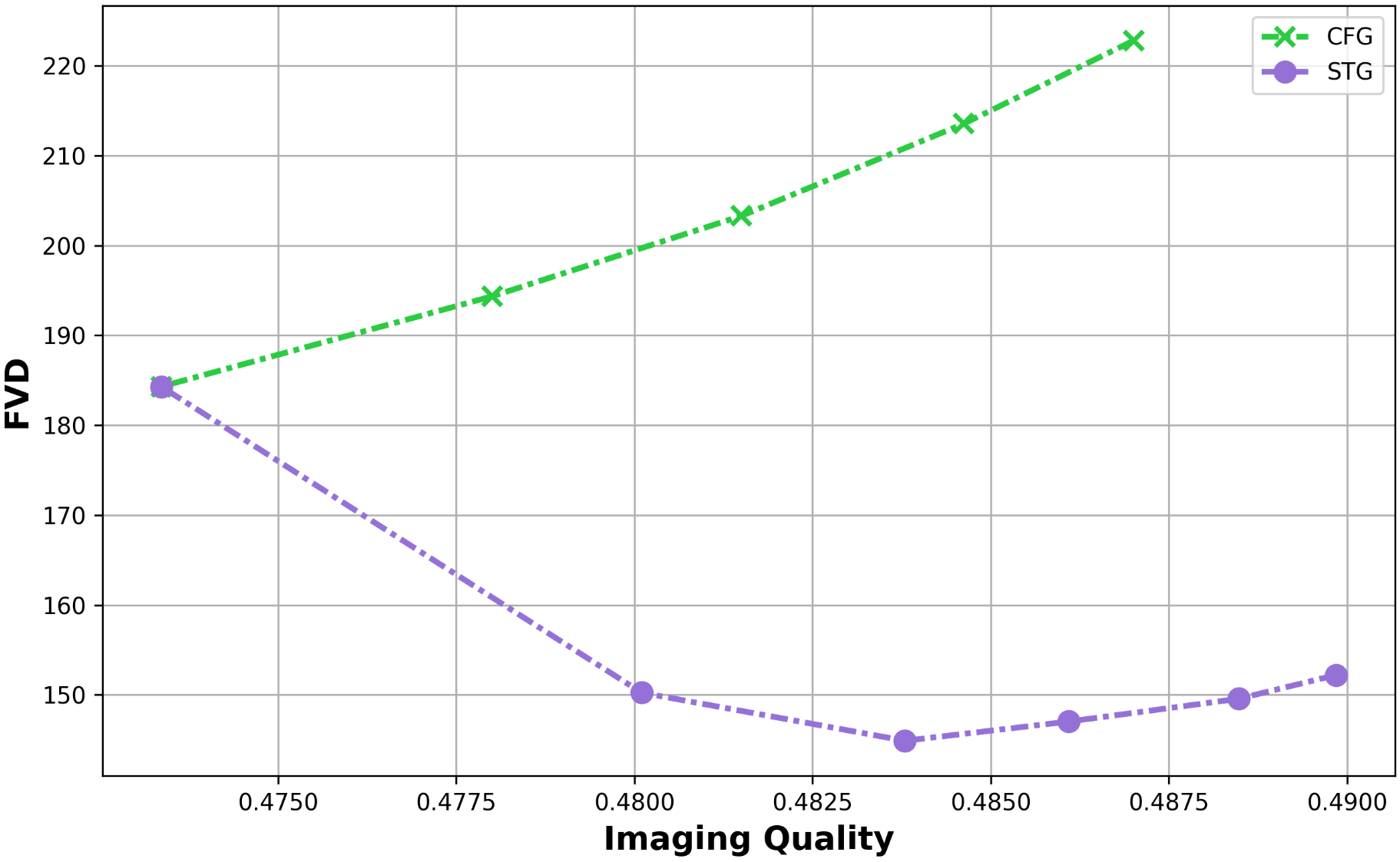}
    \caption{Comparison of CFG and STG across varying scales in terms of Imaging Quality and FVD.
    }
    \label{fig:fvd-iq_ablation}
    \vspace{-6mm}
\end{figure}

\vspace{-0.3cm}
\paragraph{Residual skip}
To skip an entire residual block, we modify $\text{Res}(z_l)$ to $\text{Res}^{'}(z_l)$,
\begin{equation}
\mathrm{Res}(z_l) =z_{l+1} = f_l(z_l) + z_l,
\label{eq:res}
\end{equation}
\begin{equation}
\mathrm{Res}^{'}(z_l) = z_{l+1} = z_l,
\label{eq:res-stg}
\end{equation}
where $z_l$ and $z_{l+1}$ represent the features at the $l^{\text{th}}$ and ${(l+1)}^{\text{th}}$ layers, respectively, and $f_l$ denotes the $l^{\text{th}}$ neural net layer.
The residual layers, which add small residuals $f_l(z_l)$ to the original feature $z_l$, ensure that the perturbed $z_{l+1}$ does not deviate significantly from the original $z_{l+1}$.
This reduces out-of-distribution issues in consecutive layers, generating perturbed but aligned samples.

\vspace{-0.3cm}

\paragraph{Attention skip}
Self-attention computes a linear combination of value tensors as
\begin{equation}
\label{eq:attention}
\mathrm{SA}(Q, K, V) = \mathrm{Softmax}\left(\frac{{Q}{K^T}}{\sqrt{d}}\right)V=\mathbf{A} V,
\end{equation}
where $ Q \in \mathbb{R}^
{(h \times w \times f) \times d}$, $K \in \mathbb{R}^{(h \times w \times f) \times d}$, $V \in \mathbb{R}^{(h \times w \times f) \times d}$ are the query, key, and value matrices, respectively. 
Here, $h$, $w$, $f$, and $d$ represent the height, width, frame number, and channel dimensions.

We can skip this layer partially by passing the value matrix directly to the next layer without computing its linear combination.
This is equivalent to replacing the attention matrix $\mathbf{A}$ with an identity matrix $\mathbf{I} \in \mathbb{R}^{hwf \times hwf}$, resulting in
\begin{equation}
\label{eq:attention_skip}
\mathrm{SA}^{'}(Q, K, V) = \mathbf{I} V.
\end{equation}
This represents a 3D extension of PAG~\cite{ahn2024self}.

\vspace{-0.3cm}
\paragraph{Factorized attention}
While recent models like Movie Gen~\cite{polyak2024movie} and Mochi~\cite{genmo2024mochi} utilize full 3D spatiotemporal attention layers, many models, such as SVD~\cite{blattmann2023stable} and Open-Sora~\cite{opensora}, still use factorized attention layers for efficiency.
These models employ sequential 2D spatial attention and 1D temporal attention to approximate 3D spatiotemporal attention. For factorized models, we apply skip perturbation of Eq.~\ref{eq:attention_skip} to spatial and temporal layers separately and use their linear combination for the final guidance.

\begin{figure*}
    \centering
    {\raggedright \small \textit{(1) A romantic scene of a couple dancing under string lights in a backyard, with warm, golden tones highlighting their laughter. \\ (2) An animation showing a floating castle drifting above the clouds, with birds flying around it and sunlight casting golden rays ... \\ (3) A realistic documentary-style video of artisans crafting pottery, with the scene unfolding and transforming as hands shape clay ... \\(4) A ghost in a white bedsheet faces a mirror. The ghost's reflection can be seen in the mirror. The ghost is in a dusty attic ... \\ }}
    \includegraphics[width=0.99\linewidth]{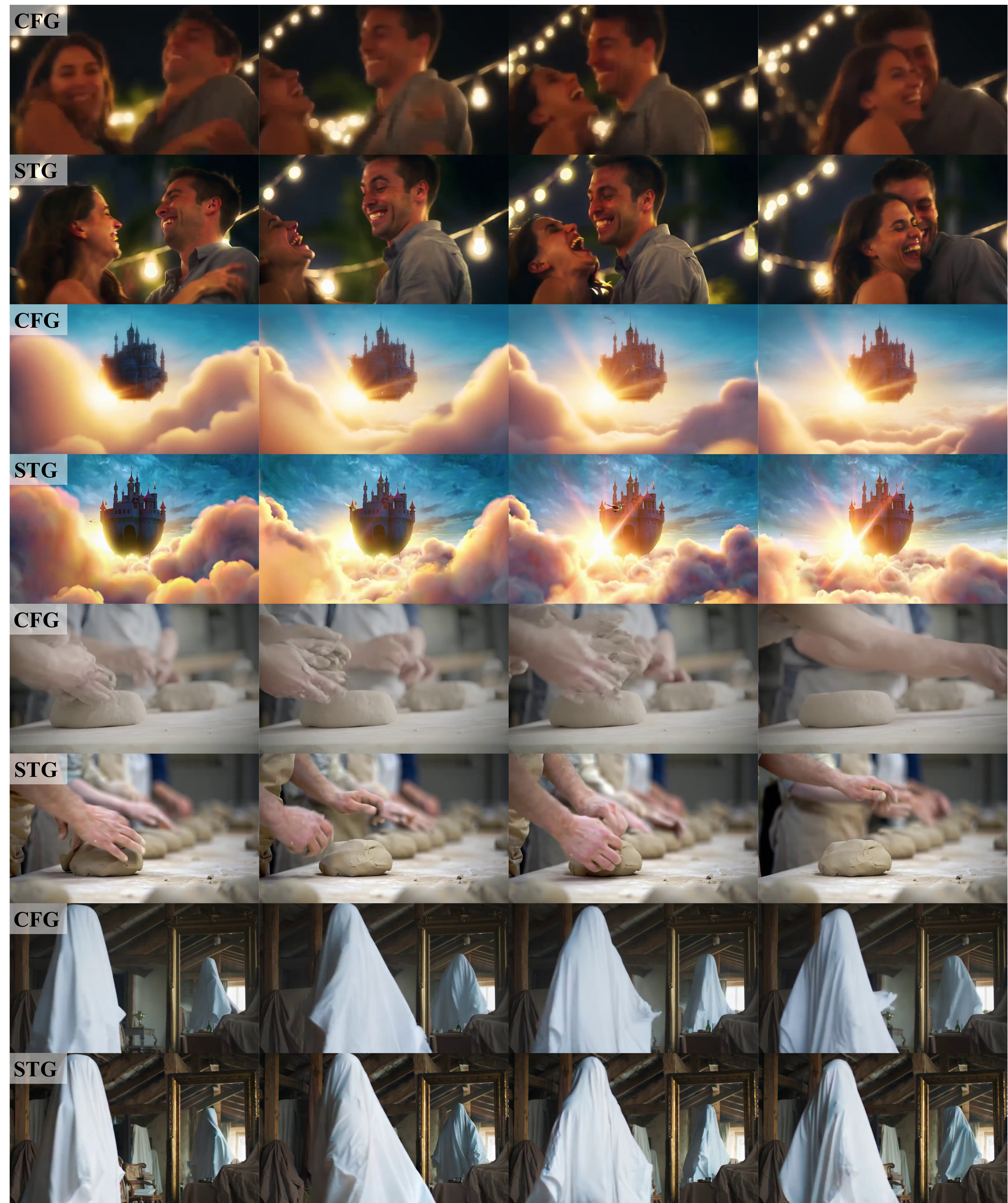}
    \caption{Qualitative comparison between CFG and STG on videos generated by Mochi~\cite{genmo2024mochi}. 
    }
    \label{fig:qual}
    \vspace{-4mm}
    
\end{figure*}

We denote the spatial and temporal perturbation labels as $y_{sb}$ and $y_{tb}$, respectively.
In practice, $y_{sb}$ and $y_{tb}$ are rarely independent, as they can influence each other.
For instance, random spatial perturbations that create varied color adjustments across frames may disrupt temporal continuity. Similarly, random temporal perturbations can affect spatial consistency, leading to distortions in specific frames. However, for our \textit{skip} perturbations, we can loosely assume independence, as skipping layers primarily reduces details in residual networks, which may not inherently affect temporal consistency.


Therefore, to simplify the derivation, we initially assume independence between spatial and temporal perturbations. 
We will revisit this assumption and propose an adjustment afterward. Under this independence assumption, the joint distribution can be expressed as
\begin{align}
    p_{\theta}(x_t | y_b) = p_{\theta}(x_t | y_{\text{sb}}) p_{\theta}(x_t | y_{\text{tb}}).
\end{align}
Following the same approach as in Eq.~\ref{eq:deriv1} and using Eq.~\ref{eq:score-ours}, we modify the score in Eq.~\ref{eq:grad_log_bayes} as follows:
\begin{align} \label{eq:score-ours-factorized}
\nabla_{x_t} \log &\Tilde{p}_{\theta}(x_t|y_g) = 
\nabla_{x_t} \log p_{\theta}(x_t) \notag \\ &+ w\nabla_{x_t} (\log p_{\theta}(x_t) - \log p_{\theta}(x_t | y_s)) \notag \\ &+ w\nabla_{x_t} (\log p_{\theta}(x_t) - \log p_{\theta}(x_t | y_t)).
\end{align}

By replacing the score with the estimated denoiser and utilizing separate scales for spatial and temporal terms, we obtain the following equation:
\begin{align}
    \tilde{\epsilon}^w_\theta(x_t) 
    &=\epsilon_\theta(x_t) + w_1 ( 
        \epsilon_\theta(x_t) - \epsilon_\theta^{\text{s}}(x_t) 
    ) \notag \\
    &\quad + w_2 ( 
        \epsilon_\theta(x_t) - \epsilon_\theta^{\text{t}}(x_t) 
    )
\label{eq:factorized-stg}
\end{align}
where $\epsilon_\theta^{\text{s}}(x_t)$ and $\epsilon_\theta^{\text{t}}(x_t)$ represent spatially and temporally perturbed models, respectively.

\begin{table*}[h]
\centering
\resizebox{\textwidth}{!}{
\begin{tabular}{lcccccccc}
\toprule
\textbf{Models} & FVD ($\downarrow$) & IS & Imaging Quality & Aesthetic Quality & Motion Smoothness & Dynamic Degree & Temporal Flickering  \\ 
\midrule
Mochi (STG-R)         &\textbf{-}   & \textbf{-}  & \textbf{0.628} & \textbf{0.554} & \textbf{0.988} & 0.86 & \textbf{0.978} \\ 
Mochi (STG-A)         & \textbf{-}  & \textbf{-}  & 0.555 & 0.541 & 0.987 & 0.86 & 0.976 \\
\midrule
Open-Sora (STG-R)         & \textbf{-}  & \textbf{-}  & 0.550 & 0.474 & 0.981 & 0.894 & \textbf{0.977}  \\ 
Open-Sora (STG-A)         &\textbf{-}   & \textbf{-}  & \textbf{0.606} & \textbf{0.509} & \textbf{0.987} & \textbf{0.895} & 0.976  \\ 
\midrule
SVD (STG-R)         & 155.9 & \textbf{39.3} & 0.687 & 0.637 & 0.965 & 0.641 & \textbf{-}    \\ 
SVD (STG-A)         & \textbf{128.7} & 38.5 & \textbf{0.694} & \textbf{0.639} & \textbf{0.968} & \textbf{0.694} & \textbf{-}   \\ 
\bottomrule
\end{tabular}
}
\caption{Comparison of STG-R (residual skip) and STG-A (attention skip) across Mochi~\cite{genmo2024mochi}, Open-Sora~\cite{opensora}, and SVD~\cite{blattmann2023stable}. STG-R shows stronger performance on Mochi, while STG-A yields better results on Open-Sora and SVD.}
\label{tab:ab_ra}
\end{table*}

\begin{table*}[h]
\centering
\resizebox{0.9\textwidth}{!}{
\begin{tabular}{lccccccc}
\toprule
\textbf{Models} & FVD ($\downarrow$) & IS & Imaging Quality & Aesthetic Quality & Motion Smoothness & Dynamic Degree  \\ 
\midrule
CFG           & 151.3  & 38.0 & 0.687                    & 0.637                    & 0.966                    & 0.562                                  \\ 
+ Spatial          & 133.8 & 38.3 & 0.691                    & \textbf{0.639}                    & 0.967                    & 0.659                                   \\ 
+ Temporal         & \textbf{128.7}  & \textbf{38.5} & \textbf{0.694}                    & 0.638                    & \textbf{0.968}                    & \textbf{0.694}                                  \\ 

\bottomrule
\end{tabular}
}
\caption{Ablation study results on SVD~\cite{blattmann2023stable} factorized attention, showing the impact of adding spatial and temporal guidance.}
\vspace{-2mm}
\label{tab:ab_st}
\end{table*}

Next, we revisit our assumption of independence between spatial and temporal perturbations. While Eq.~\ref{eq:factorized-stg} works well in practice, we can derive an alternative STG formulation that uses orthogonalization to isolate the independent components of the spatial and temporal guidance. Inspired by a negative prompting technique~\cite{armandpour2023re}, we can modify the Eq.~\ref{eq:factorized-stg} as follows:
\begin{multline}
    \tilde{\epsilon}^w_\theta(x_t) = \epsilon_\theta(x_t) + w_1 \Delta_\text{s} \\ +  w_2 \left(\Delta_\text{t}
    - \frac{\langle \Delta_\text{s}, \Delta_\text{t} \rangle}{\| \Delta_\text{s} \|^2} \, \Delta_\text{s}\right),
\end{multline}
where $\Delta_\text{s}= \epsilon_\theta(x_t) - \epsilon_\theta^{\text{s}}(x_t)$ and $\Delta_\text{t} = \epsilon_\theta(x_t) - \epsilon_\theta^{\text{t}}(x_t)$.

\paragraph{Manifold Constrained Guidance} 

Even with a well-aligned weak model, larger guidance scales $w$ can drive samples off the data manifold, resulting in poor quality and oversaturation. To address error accumulation at high guidance scales, we explore optional techniques to keep samples constrained to the manifold. 
Rescaling the latent code~\cite{lin2024common} helps mitigate this issue by constraining its variance, as larger variance is known to cause saturation in the results. Additionally, Restart sampling~\cite{xu2023restart} demonstrates that introducing stochasticity can correct off-manifold deviations. Building on this idea, we incorporate stochastic forward processes into our sampling guidance framework as an optional method. While this approach modestly improves final sample quality and reduces saturation, it introduces additional computational overhead. Further details are provided in Appendix~\ref{subsec:supple_mcg}.

\section{Experiments}
\label{sec:experiments}

\subsection{Overview}
We employ three models for our experiments:

\begin{itemize}
    \item \textbf{Mochi}~\cite{genmo2024mochi} is a text-to-video model built on AsymmDiT blocks, containing a total of 10 billion parameters and utilizing 3D self-attention in its spatiotemporal layers.
    \item \textbf{Open-Sora}~\cite{opensora} is a text-to-video model built on STDiT blocks with 1.1 billion parameters, employing factorized spatial and temporal attention layers.
    \item \textbf{SVD}~\cite{blattmann2023stable} is an image-to-video model with 1.5 billion parameters that leverages factorized spatial and temporal attention within a UNet architecture.
\end{itemize}

We evaluate the proposed method using widely adopted datasets and metrics to ensure a comprehensive performance analysis.




\begin{itemize}
    \item \textbf{UCF-101} The UCF-101 dataset comprises 13,320 videos organized into 101 action classes. Using this dataset, we assess the Fréchet Video Distance (FVD) and Inception Score (IS) of the image-to-video model SVD~\cite{blattmann2023stable}. Following DIGAN~\cite{yu2022generating}, we calculate FVD and IS on 2,048 and 10,000 samples, respectively. For conditioning, initial frames from UCF-101 videos are used as inputs to the SVD model, with each input frame serving as the starting frame of the generated videos.

    \item \textbf{VBench} We use VBench~\cite{huang2024vbench} for automatic evaluation across various video metrics. SVD is evaluated using the image-to-video (I2V) framework on 355 samples from the VBench I2V dataset with 5 random seeds. Mochi and Open-Sora are evaluated with the text-to-video (T2V) framework: Open-Sora uses the standard VBench prompt list, while Mochi uses 100 randomly selected prompts due to computational limits.

    \item \textbf{EvalCrafter} We perform human evaluations using 700 prompts from the EvalCrafter dataset~\cite{liu2024evalcrafter}.

    \item \textbf{LLM-Generated Prompts} We use a set of 100 selected prompts generated by Claude 3.5 Sonnet~\cite{TheC3} for our demo and qualitative comparisons.
\end{itemize}


For evaluation, CFG is used together with STG for our evaluation. We did not use Restart sampling or guidance orthogonalization; additional results with these techniques are in the Appendix sections~\ref{subsec:supple_mcg} and~\ref{subsec:supple_ortho}.

\subsection{Results}

\noindent\textbf{Qualitative Comparison}
Fig.~\ref{fig:qual} compares CFG and our method. Videos from the naive CFG model often show blurry objects with indistinct shapes, while STG produces clearer, more vivid frames with sharper image quality. STG also reduces temporal inconsistency and flickering, especially in dynamic videos with large motion where CFG frequently fails. Spatial guidance enhances object structure, and temporal guidance improves consistency. Please refer to the Appendix sections~\ref{sec:supple_ab} and~\ref{sec:supple_qual} for additional results.


\noindent\textbf{Quantitative Comparison}
We compare CFG and STG using the FVD-Imaging Quality (VBench) plot across different scales in Fig.~\ref{fig:fvd-iq_ablation}. 
FVD measures video distribution, while Imaging Quality assesses frame clarity. Higher CFG scales improve Imaging Quality but reduce diversity, as reflected in higher FVD. STG avoids this trade-off, maintaining diversity at increased scales.

T2V and I2V VBench metrics in Tab.\ref{tab:quant-t2v} and Tab.\ref{tab:quant-i2v} show notable improvements in frame-level quality (Imaging and Aesthetic Quality). Temporal quality improves qualitatively, though metrics like Motion Smoothness and Temporal Flickering show marginal gains, as these scores are near saturation $(\sim 0.9\text{x})$.


For Dynamic Degree, we aim to keep it unchanged, as it may not correlate with video quality. This is achieved in the T2V metric, but in the I2V model, CFG increases the influence of the conditioned image, reducing motion. STG, while not directly affecting Dynamic Degree, mitigates CFG's impact, leading to increased Dynamic Degree in I2V when used together.

\noindent\textbf{Human Evaluation}
We provide human evaluation results on 700 prompts from the EvalCrafter dataset~\cite{liu2024evalcrafter} in the Appendix~\ref{sec:supple_humaneval}.

\subsection{Ablation Study}
Fig.~\ref{fig:ablation} displays selected frames from videos generated by Mochi using various STG scales, where \( w=0 \) corresponds to the CFG-only model without STG.
Increasing the STG scale results in more vivid colors and finer details compared to the monotonous colors at $w=0$.
Notably, unlike with CFG, sampling diversity is preserved as the STG scale increases, as shown in Fig.~\ref{fig:fvd-iq_ablation}.

We performed an ablation study on SVD to evaluate spatial and temporal guidance (Tab.~\ref{tab:ab_st}). Spatial guidance notably reduced FVD and improved metrics, while temporal guidance further boosted performance, confirming their combined effectiveness for spatiotemporal generation.

We also test two STG variants—residual skip (STG-R) and attention skip (STG-A)-on SVD, Open-Sora, and Mochi. 
As shown in Tab.~\ref{tab:ab_ra}, STG-R performs well with Mochi, while STG-A is more effective for the other models.
This is likely due to Mochi's higher parameter count and layer depth, enabling more extensive skipping without triggering out-of-distribution (OOD) issues in consecutive layers. Furthermore, Mochi’s spatiotemporal layers consist of a single 3D attention layer, so STG-R skips a residual layer with just one attention layer. In contrast, SVD and Open-Sora use factorized attention, meaning STG-R skips a residual layer with two attention layers, which may cause excessive perturbation and lead to OOD issues in subsequent layers.
More results are available in the Appendix~\ref{sec:supple_ab}.

\vspace{-0.6mm}

\section{Conclusion}
\label{sec:conclusion}

We proposed Spatiotemporal Skip Guidance (STG), a simple, training-free method for video diffusion models. By simulating an aligned weak model via spatiotemporal skipping, STG offers strong guidance for high-fidelity video generation, achieving notable qualitative and quantitative improvements.
We hope this work advances video diffusion models and inspires further research in the field.

\paragraph{Limitation and Ethical Considerations}

STG's performance depends on scale and layer selection, with the optimal configuration varying across models, requiring users to set these through heuristic tuning. While video quality improvements are notable, they also raise ethical concerns about misuse, underscoring the importance of using this technology responsibly and constructively.

{
    \small
    \bibliographystyle{ieeenat_fullname}
    \bibliography{main}
}


\clearpage

\setcounter{page}{1}


\begin{onecolumn}













\renewcommand{\thesection}{A\arabic{section}}
\setcounter{section}{0} 

\section{Experimental Details}
\subsection{Sampling Algorithm}
\begin{algorithm}[htbp]
\caption{Spatiotemporal Skip Guidance (STG)}
\KwIn{$\epsilon_\theta$, $\epsilon_\theta^{s,t}$: Main model and spatiotemporally perturbed model respectively. \\
$w$: Spatiotemporal guidance scale. \\
$\Sigma_t$: Variance at step $t$.}
\KwOut{Generated video $V_{\text{out}}$.}

$x_T \sim \mathcal{N}(0, I)$\

\For{$t \gets T, T-1, \dots, 1$}{
    $\epsilon_t \gets \epsilon_\theta(x_t)$ \
    $\epsilon_t^{s,t} \gets \epsilon_\theta^{s,t}(x_t)$ \
    $\tilde{\epsilon}_t \gets \epsilon_t + w (\epsilon_t - \epsilon_t^{s,t})$ \
    $x_{t-1} \sim \mathcal{N}\left(\frac{1}{\sqrt{\alpha_t}} \left( x_t - \frac{1-\alpha_t}{\sqrt{1-\bar{\alpha}_t}} \tilde{\epsilon}_t \right), \Sigma_t\right)$ \
}

\Return{$V_{\text{out}}$}\
\end{algorithm}
\vspace{-4mm}

\begin{algorithm}[htbp]
\caption{Spatiotemporal Skip Guidance (STG) for factorized attention}
\KwIn{$\epsilon_\theta$, $\epsilon_\theta^s$, $\epsilon_\theta^t$: Main model, spatially perturbed, and temporally perturbed models respectively. \\
$w_1$, $w_2$: Guidance scales. \\
$\Sigma_t$: Variance at step $t$.}
\KwOut{Generated video $V_{\text{out}}$.}

$x_T \sim \mathcal{N}(0, I)$\

\For{$t \gets T, T-1, \dots, 1$}{
    $\epsilon_t \gets \epsilon_\theta(x_t)$ \
    $\epsilon_t^s \gets \epsilon_\theta^s(x_t)$ \
    $\epsilon_t^t \gets \epsilon_\theta^t(x_t)$ \
    $\tilde{\epsilon}_t \gets \epsilon_t + w_1 (\epsilon_t - \epsilon_t^s) + w_2 (\epsilon_t - \epsilon_t^t)$ \
    $x_{t-1} \sim \mathcal{N}\left(\frac{1}{\sqrt{\alpha_t}} \left( x_t - \frac{1-\alpha_t}{\sqrt{1-\bar{\alpha}_t}} \tilde{\epsilon}_t \right), \Sigma_t\right)$ \
}

\Return{$V_{\text{out}}$}\
\end{algorithm}

\vspace{-4mm}

\subsection{Computational Resources}
For evaluation, we utilized an NVIDIA A100 40GB GPU for SVD~\cite{blattmann2023stable}, while Open-Sora~\cite{opensora} and Mochi~\cite{genmo2024mochi} were evaluated using NVIDIA H100 or A100 80GB GPUs.

\subsection{Implementation Details}
The default model scales for CFG are as follows: SVD~\cite{blattmann2023stable} uses a scale of 3.0, Open-Sora~\cite{opensora} uses 7.0, and Mochi~\cite{genmo2024mochi} uses 4.5. For STG, the configurations vary, with STG-A using a scale of 2.0 and STG-R using 1.0.
STG is applied to the $8^{\text{th}}$ layer of SVD, which has a total of 16 layers, and the $12^{\text{th}}$ layer of Open-Sora, which has a total of 28 layers. For Mochi, which has 48 layers in total, STG is applied at the $35^{\text{th}}$ layer.


\subsection{Metrics}
   
To evaluate model performance across different datasets, several methodologies were employed. FVD was assessed using the VideoMAE~\cite{ge2024content} model. IS was evaluated with the C3D model~\cite{tran2015learning, jia2014caffe, karpathy2014large}, following the setup of TGAN~\cite{ding2019tgan}. For VBench - Imaging Quality, the MUSIQ image quality predictor, trained on the SPAQ dataset, was used. VBench - Aesthetic Quality was measured using the LAION aesthetic predictor, applied to individual video frames. VBench - Dynamic Degree was evaluated with the RAFT flow estimator to quantify the degree of dynamics. VBench - Motion Smoothness was calculated as the mean absolute error (MAE) between dropped and reconstructed frames using a video frame interpolation model. Finally, VBench - Temporal Flickering was assessed by generating static frames and computing the mean absolute difference between consecutive frames.

\subsection{Prompts Used}


\begin{tcolorbox}[mybox, title=EvalCrafter prompt, center title]
\textit{
\begin{enumerate}
    \item 2 Dog and a whale, ocean adventure
    \item Teddy bear and 3 real bear
    \item Goldfish in glass
    \item A small bird sits atop a blooming flower stem.
    \item A fluffy teddy bear sits on a bed of soft pillows surrounded by children's toys.
    \item A peaceful cow grazing in a green field under the clear blue sky.
    \item Unicorn sliding on a rainbow
    \item Four godzillas
    \item A fluffy grey and white cat is lazily stretched out on a sunny window sill, enjoying a nap after a long day of lounging.
    \item A curious cat peers from the window, watching the world outside.
    \item A horse
    \item A pig
    \item A squirrel
    \item A bird
    \item A zebra
    \item Two elephants are playing on the beach and enjoying a delicious beef stroganoff meal.
    \item Two fish eating spaghetti on a subway
    \item A pod of dolphins gracefully swim and jump in the ocean.
    \item A peaceful cow grazing in a green field under the clear blue sky.
    \item A cute and chubby giant panda is enjoying a bamboo meal in a lush forest. The panda is relaxed and content as it eats, and occasionally stops to scratch its ear with its paw.
    \item Dragon flying over the city at night
    \item Pikachu snowboarding
    \item A cat drinking beer
    \item A dog wearing VR goggles on a boat
    \item A giraffe eating an apple
    \item Five camels walking in the desert
    \item Mickey Mouse is dancing on white background
    \item A happy pig rolling in the mud on a sunny day.
    \item In an African savanna, a majestic lion is prancing behind a small timid rabbit. The rabbit tried to run away, but the lion catches up easily.
    \item 3 sheep enjoying spaghetti together
    \item A photo of a Corgi dog riding a bike in Times Square. It is wearing sunglasses and a beach hat.
    \item A pod of dolphins gracefully swim and jump in the ocean.
    \item In the lush forest, a tiger is wandering around with a vigilant gaze while the birds chirp and monkeys play.
    \item The teddy bear and rabbit were snuggled up together. The teddy bear was hugging the rabbit, and the rabbit was nuzzled up to the teddy bear's soft fur.
    \item A slithering snake moves through the lush green grass.
    \item A pair of bright green tree frogs cling to a branch in a vibrant tropical rainforest.
    \item Four fluffy white Persian kittens snuggle together in a cozy basket by the fireplace.
    \item Eight fluffy yellow ducklings waddle behind their mother, exploring the edge of a pond.
    \item A family of four fluffy, blue penguins waddled along the icy shore.
    \item Two white swans gracefully swam in the serene lake.
    \item In a small forest, a colorful bird was flying around gracefully. Its shiny feathers reflected the sun rays, creating a beautiful sight.
    \item A spider spins a web, weaving intricate patterns with its silk.
    \item ...
\end{enumerate}
}


\end{tcolorbox}

\clearpage
\begin{tcolorbox}[mybox, title=Curated prompt for demos, center title]
\textit{
\begin{enumerate}
    \item Sloth with pink sunglasses lays on a donut float in a pool. The sloth is holding a tropical drink. The world is tropical. The sunlight casts a shadow.
    \item A vibrant, top-down video of a kayak gliding through multicolored waters, showcasing shifting hues from blue to red to illustrate varying flow speeds or temperatures. The paddle interacts with the water, creating dynamic ripples and currents.
    \item A slow-motion capture of a beautiful woman in a flowing dress spinning in a field of sunflowers, with petals swirling around her.
    \item An exotic video of rabbits on the moon making rice cakes under a star-filled sky, with Earth visible in the background.
    \item A handsome man walking confidently through a bustling city street at night, illuminated by neon lights and reflections in puddles.
    \item A close-up shot of a butterfly landing on the nose of a woman, highlighting her smile and the intricate details of the butterfly's wings.
    \item A top-down video of a table filled with colorful dishes from different cuisines, with hands reaching in to serve food and clinking glasses.
    \item A majestic bird’s-eye view of a couple holding hands while walking along the shore of a beach with sparkling turquoise waves.
    \item A surreal scene of a forest where the leaves glow in neon colors, with a person walking down a path as fireflies dance around them.
    \item A drone shot of a desert at sunset, where shadows stretch and shift, capturing a lone traveler moving gracefully through the sand dunes.
    \item A close-up of a beautiful woman's face with colored powder exploding around her, creating an abstract splash of vibrant hues.
    \item A panoramic view of a tropical waterfall surrounded by lush greenery, with a rainbow forming in the mist.
    \item A whimsical video of floating lanterns being released into the sky over a calm lake, with reflections on the water creating a mirror effect.
    \item A time-lapse video of an artist painting a mural on a city wall, where each frame shows a burst of color and detail.
    \item An overhead video of koi fish swimming in a pond with rippling water, with their scales reflecting shades of gold, orange, and white.
    \item A slow-motion clip of a handsome person diving into a crystal-clear ocean, with water splashing and bubbles forming intricate patterns.
    \item A fantastical scene of a meadow where flowers bloom and change colors in sync with the music, with a person dancing among them.
    \item A top-down video of a hot air balloon festival, showing multicolored balloons lifting off and dotting the sky.
    \item A beautiful woman sitting by a window as rain drizzles down, creating streaks and patterns on the glass.
    \item A cinematic shot of a person walking through a field of lavender during golden hour, with the wind gently swaying the purple blossoms.
    \item An exotic video of floating jellyfish in the ocean, their translucent bodies glowing with bioluminescence in shades of blue and purple.
    \item A playful video of puppies running across a vibrant, flower-filled meadow, filmed in slow motion to capture their joyful expressions.
    \item A captivating aerial view of a cityscape at sunrise, with skyscrapers casting long shadows and golden light reflecting on windows.
    \item A stunning slow-motion shot of a bird taking flight over a reflective lake, with water droplets glistening as they scatter.
    \item A romantic scene of a couple dancing under string lights in a backyard, with warm, golden tones highlighting their laughter.
    \item ...
\end{enumerate}
}



\end{tcolorbox}

\begin{tcolorbox}[mybox, title=Prompt for figures in the main paper, center title]
\textit{
    \textbf{Fig2:} 
    \begin{itemize}
    \item A macro cinematography animation showing a butterfly emerging from its chrysalis, filmed with side-lit lighting to accentuate the texture of its wings.
    \end{itemize}
    \textbf{Fig4:} 
    \begin{itemize}
        \item A 50mm lens shot of a couple embracing under string lights as the camera slowly tracks them, capturing their shared laughter in a soft, cinematic glow.
        \item An animation showing a floating castle drifting above the clouds, with birds flying around it and sunlight casting golden rays, evoking the feeling of wonder seen in classic animations.
        \item A realistic documentary-style video of artisans crafting pottery, with the scene unfolding and transforming as hands shape clay under diffused lighting.
        \item A ghost in a white bedsheet faces a mirror. The ghost's reflection can be seen in the mirror. The ghost is in a dusty attic, filled with old beams, cloth-covered furniture. The attic is reflected in the mirror. The light is cool and natural. The ghost dances in front of the mirror.
    \end{itemize}
}
\end{tcolorbox}

\section{Human Evaluation}
\label{sec:supple_humaneval}
We conducted user studies following EvalCrafter~\cite{liu2024evalcrafter} to evaluate subjective opinions across five key aspects: (1) Video Quality, reflecting the clarity of the generated video, with higher scores indicating reduced blur, noise, or visual artifacts; (2) Text-Video Alignment, assessing the correspondence between the input text prompt and the generated video, particularly focusing on the accuracy of generated motions; (3) Motion Quality, evaluating the correctness and realism of the motions depicted in the video; (4) Temporal Consistency, measuring the frame-to-frame coherence, distinct from Motion Quality as it requires users to assess the smoothness of movement; and (5) Subjective Likeness, akin to an aesthetic score, where higher values signify better alignment with human preferences.

For each metric, feedback was collected from seven users, who rated videos on a scale from 1 to 5, with higher scores representing better alignment. To ensure fairness, the video sequences were randomly shuffled before being presented to users.

We used 700 prompts from EvalCrafter for text-to-video (T2V) generation with Mochi~\cite{genmo2024mochi}. Additionally, we employed FLUX.1 [dev]~\cite{flux_repo} to generate images from these prompts, which served as input to the image-to-video (I2V) model (SVD~\cite{blattmann2023stable}). The results, shown in Fig.~\ref{fig:user_study}, demonstrate that incorporating STG leads to improved quality across all evaluated aspects.

\begin{figure*}[b]
    \centering
    \includegraphics[width=\linewidth]{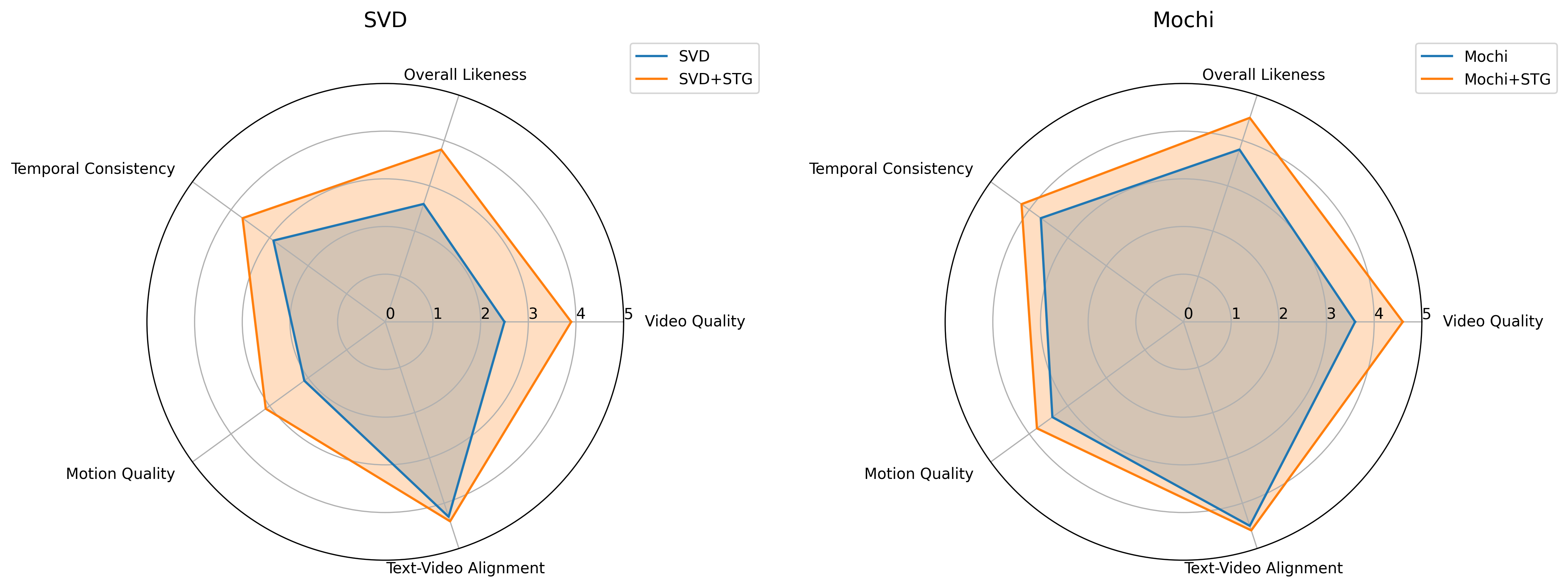}
    \caption{User study results for STG on SVD and Mochi, using 700 prompts from EvalCrafter~\cite{liu2024evalcrafter}. For I2V generation of SVD, we employed FLUX.1 [dev]~\cite{flux_repo} to generate images from these prompts, which served as input to the model. The results demonstrate that incorporating STG leads to improved quality across all evaluated aspects.
    }
    \label{fig:user_study}
    \vspace{-2mm}
\end{figure*}

\section{Ablation Study}
\label{sec:supple_ab}

\subsection{Manifold Constrained Guidance}
\label{subsec:supple_mcg}

As discussed in the main paper, sampling guidance techniques, including STG, utilize scale guidance, which can sometimes cause the sampling trajectory to deviate from the data manifold. This deviation is particularly noticeable when STG is applied with large scales or to videos that are already bright, often resulting in broken videos or over-saturation due to manifold overshooting. To mitigate these issues, we propose a set of optional techniques that can serve as effective remedies.

First, we leverage the error contraction property of stochastic processes~\cite{xu2023restart} by incorporating stochastic forward processes into the sampling guidance framework. This technique, referred to as \textbf{STG with Restart}, is detailed in Algorithm~\ref{algo:restart}. While this method moderately enhances the quality of the final samples and resolves issues such as broken videos (as illustrated in Fig.~\ref{fig:restart_stg}), it introduces additional computational overhead.

Additionally, increased variance in the latent code~\cite{lin2024common} has been observed in over-saturated results. Consequently, over-saturation can be effectively mitigated using a rescaling technique~\cite{lin2024common}, which constrains the variance of the latent code. This method, referred to as \textbf{STG with Rescaling}, is detailed in Algorithm~\ref{algo:rescale}. As shown in Fig.~\ref{fig:rescale_stg}, videos generated with larger variance (second row) often display saturated colors, which are successfully resolved by applying variance rescaling (third row). Unlike the Restart method, Rescaling introduces negligible computational overhead, making it the preferred approach for addressing over-saturation.

\begin{algorithm}[htbp]
\caption{Spatiotemporal Skip Guidance with Restart}
\KwIn{$\epsilon_\theta$, $\epsilon_\theta^{s,t}$: Main model and spatiotemporal perturbed model respectively. \\
$w$: Spatiotemporal guidance scale. \\
$\Sigma_t$: Variance at step $t$. \\
$t_{\text{min}}, t_{\text{max}}$: Restart interval. \\
$K$: Number of Restart iterations.}
\KwOut{Generated video $V_{\text{out}}$.}

$x_T \sim \mathcal{N}(0, I)$\

\For{$t \gets T, T-1, \dots, 1$}{
    $\epsilon_t \gets \epsilon_\theta(x_t)$\
    $\epsilon_t^{s,t} \gets \epsilon_\theta^{s,t}(x_t)$\
    $\tilde{\epsilon}_t \gets \epsilon_t + w (\epsilon_t - \epsilon_t^{s,t})$\
    $x_{t-1} \sim \mathcal{N}\left(\frac{1}{\sqrt{\alpha_t}} \left( x_t - \frac{1-\alpha_t}{\sqrt{1-\bar{\alpha}_t}} \tilde{\epsilon}_t \right), \Sigma_t\right)$\

    \If{$t = t_{\text{min}}$}{
        $x_{t_{\text{min}}}^0 \gets x_{t-1}$\

        \For{$k \gets 0, \dots, K-1$}{
            $\epsilon_{\text{restart}} \sim \mathcal{N}(0, \Sigma_{\text{restart}})$\
            $x_{t_{\text{max}}}^{k+1} \gets x_{t_{\text{min}}}^k + \epsilon_{\text{restart}}$\

            \For{$t' \gets t_{\text{max}}, t_{\text{max}}-1, \dots, t_{\text{min}}$}{
                $\epsilon_{t'} \gets \epsilon_\theta(x_{t'}^{k+1})$\
                $\epsilon_{t'}^{s,t} \gets \epsilon_\theta^{s,t}(x_{t'}^{k+1})$\
                $\tilde{\epsilon}_{t'} \gets \epsilon_{t'} + w (\epsilon_{t'} - \epsilon_{t'}^{s,t})$\
                $x_{t'-1}^{k+1} \sim \mathcal{N}\left(\frac{1}{\sqrt{\alpha_{t'}}} \left( x_{t'}^{k+1} - \frac{1-\alpha_{t'}}{\sqrt{1-\bar{\alpha}_{t'}}} \tilde{\epsilon}_{t'} \right), \Sigma_{t'}\right)$\
            }
        }
    }
}

\Return{$V_{\text{out}}$}\
\label{algo:restart}
\end{algorithm}

\begin{algorithm}[htbp]
\caption{Spatiotemporal Skip Guidance (STG) with Rescaling}
\KwIn{$\epsilon_\theta$, $\epsilon_\theta^{s,t}$: Main model and spatiotemporal perturbed model respectively. \\
$w$: Spatiotemporal guidance scale. \\
$rescale$: Rescaling factor. \\
$\Sigma_t$: Variance at step $t$.}
\KwOut{Generated video $V_{\text{out}}$.}

$x_T \sim \mathcal{N}(0, I)$\

\For{$t \gets T, T-1, \dots, 1$}{
    $\epsilon_t \gets \epsilon_\theta(x_t)$ \
    $\epsilon_t^{s,t} \gets \epsilon_\theta^{s,t}(x_t)$ \
    $\tilde{\epsilon}_t \gets \epsilon_t + w (\epsilon_t - \epsilon_t^{s,t})$ \
    
    $\text{std}_{\epsilon} \gets \text{std}(\epsilon_t)$ \
    $\text{std}_{\tilde{\epsilon}} \gets \text{std}(\tilde{\epsilon}_t)$ \
    $\text{factor} \gets \frac{\text{std}_{\epsilon}}{\text{std}_{\tilde{\epsilon}}}$ \
    $\text{factor} \gets rescale \cdot \text{factor} + (1 - rescale)$ \
    $\tilde{\epsilon}_t \gets \tilde{\epsilon}_t \cdot \text{factor}$ \
    
    $x_{t-1} \sim \mathcal{N}\left(\frac{1}{\sqrt{\alpha_t}} \left( x_t - \frac{1-\alpha_t}{\sqrt{1-\bar{\alpha}_t}} \tilde{\epsilon}_t \right), \Sigma_t\right)$ \
}

\Return{$V_{\text{out}}$}\
\label{algo:rescale}
\end{algorithm}

\begin{table*}
\centering
\resizebox{0.9\textwidth}{!}{%
\begin{tabular}{lcccccc}
\toprule
\textbf{Models}    & FVD ($\downarrow$) & IS           & Imaging Quality & Aesthetic Quality & Motion Smoothness & Dynamic Degree  \\
\midrule

SVD (STG) & \textbf{128.7} & \textbf{38.5}& \textbf{0.694} & \textbf{0.639} & \textbf{0.968} & \textbf{0.694} \\
SVD (STG-ORTH) & 130.4 & 38.4& 0.691 & 0.637 & 0.967 & 0.692  \\
\bottomrule
\end{tabular}
}
\caption{Ablation results of STG on SVD~\cite{blattmann2023stable}, evaluating the impact of orthogonalizing spatial and temporal guidance (STG-ORTH). Our findings show no performance gain from applying orthogonalization; therefore, we do not adopt it.}
\label{tab:orthogonal}
\end{table*}

\subsection{STG with Orthogonalization}
\label{subsec:supple_ortho}
As discussed in the main paper, for SVD and Open-Sora, which utilize factorized spatial and temporal attention, it is possible to orthogonalize spatial and temporal guidance. The detailed algorithm for this approach is provided in Algorithm~\ref{algo:ortho}. However, we do not implement orthogonalization in practice, as it does not demonstrate any performance improvement, as shown in Table~\ref{tab:orthogonal}.

\begin{algorithm}[htbp]
\caption{Spatiotemporal Skip Guidance (STG) with Orthogonalization}
\KwIn{$\epsilon_\theta$, $\epsilon_\theta^s$, $\epsilon_\theta^t$: Main model, spatially perturbed, and temporally perturbed models respectively. \\
$w_1$, $w_2$: Guidance scales. \\
$\Sigma_t$: Variance at step $t$.}
\KwOut{Generated video $V_{\text{out}}$.}

$x_T \sim \mathcal{N}(0, I)$\

\For{$t \gets T, T-1, \dots, 1$}{
    $\epsilon_t \gets \epsilon_\theta(x_t)$ \
    $\epsilon_t^s \gets \epsilon_\theta^s(x_t)$ \
    $\epsilon_t^t \gets \epsilon_\theta^t(x_t)$ \

    $\Delta_s \gets \epsilon_t - \epsilon_t^s$ \
    $\Delta_t \gets \epsilon_t - \epsilon_t^t$ \

    $\Delta_t^{\perp} \gets \Delta_t - \frac{\langle \Delta_s, \Delta_t \rangle}{\|\Delta_s\|^2} \cdot \Delta_s$ \

    $\tilde{\epsilon}_t \gets \epsilon_t + w_1 \Delta_s + w_2 \Delta_t^{\perp}$ \

    $x_{t-1} \sim \mathcal{N}\left(\frac{1}{\sqrt{\alpha_t}} \left( x_t - \frac{1-\alpha_t}{\sqrt{1-\bar{\alpha}_t}} \tilde{\epsilon}_t \right), \Sigma_t\right)$ \
}

\Return{$V_{\text{out}}$}\
\label{algo:ortho}
\end{algorithm}

\subsection{Layer Ablation}
STG can be applied to different layers, and we conduct an ablation study to evaluate the impact of skipping various layers for STG on Mochi~\cite{genmo2024mochi}. The results are presented in Fig.~\ref{fig:layer-ab}. Mochi consists of 48 layers in total, and we experimented with layer skipping at layers 30, 32, and 35. Our findings show that skipping later layers has a more significant effect on quality improvements, as these layers are primarily responsible for refining texture details. Throughout all experiments in this paper, we consistently skip layer 35.

\subsection{Effect of Spatial and Temporal Guidance}
For models with factorized attention layers, guidance can be applied separately to spatial and temporal layers. When using STG-A, it functions similarly to applying PAG~\cite{ahn2024self} to the spatial attention layers, and we refer to this method as Spatial PAG (SPAG).
When spatial guidance is applied alone, as shown for SVD in Fig.~\ref{fig:cfg_pag_stg_svd} and for Open-Sora in Fig.~\ref{fig:cfg_pag_stg_opensora}, the results struggle to maintain clarity during motion and exhibit poor temporal consistency. For instance, significant artifacts appear near the wings in the second row of Fig.~\ref{fig:cfg_pag_stg_svd}, and around the legs of the chicken in Fig.~\ref{fig:cfg_pag_stg_opensora}.

We further investigate the individual contributions of spatial and temporal guidance. In Fig.~\ref{fig:spag}, we compare results with and without Spatial Guidance (SPAG). The results show that while CFG fails to maintain clear object structures, resulting in blurry videos, SPAG significantly enhances object structure and improves clarity.

Similarly, in Fig.~\ref{fig:tpag}, we present results with and without Temporal Guidance (TPAG). The results reveal that CFG struggles to ensure frame-to-frame consistency, with the shape and color of the jelly varying noticeably across frames, leading to a disjointed video. In contrast, TPAG effectively preserves the jelly's appearance throughout the sequence, creating a more cohesive video and significantly improving Temporal Consistency.

\subsection{Attention Skip and Residual Skip}
We compare the performance of attention skip (STG-A) and residual skip (STG-R) in Mochi~\cite{genmo2024mochi} and Open-Sora~\cite{opensora}. The results for Mochi in Fig.~\ref{fig:stgar_mochi} indicate that STG-R delivers greater qualitative improvements for Mochi.
On the other hand, the results for Open-Sora in Fig.~\ref{fig:stgar_opensora} and SVD in Fig.~\ref{fig:stgar_svd} demonstrate that STG-A delivers greater qualitative improvements for these models.
Based on these findings, we use STG-A for Open-Sora and SVD, and use STG-R for Mochi in all experiments presented in the paper.

\subsection{Weak Model Visualization}
We visualize the results of one-step prediction using different denoising methods (weak models, CFG, and STG) at timestep 30 in Fig.~\ref{fig:weak_t30} and timestep 24 in Fig.~\ref{fig:weak_t24}, rows (a) to (e).
Row (c) shows results denoised using the spatiotemporally perturbed model, $\epsilon^{s,t}_{\theta}(x_t)$, which generally produces blurrier outcomes compared to row (b), where the unconditional weak model $\epsilon_{\theta}(x_t | \phi)$ of CFG is applied. By moving away from the blurry weak model, STG achieves clear and well-defined structures with natural color tones. In contrast, CFG often produces unnatural color artifacts and broken structures. For example, the video predicted by CFG renders the girl’s arm on the left unnaturally red, the man’s arm on the right unnaturally dark, and the trees and leaves in the background blurry. By comparison, STG consistently generates videos with enhanced structure and natural, well-balanced color tones.

\subsection{Other Perturbation Methods}
In addition to SPAG (spatial perturbation using PAG), we explore other perturbation techniques. One such approach is SEG~\cite{hong2024smoothed}, which applies Gaussian blurring to the attention map.
A comparison of CFG, SEG, and STG is presented in Fig.~\ref{fig:seg_comp}. The results frequently show broken outputs in both CFG and SEG. In contrast, incorporating layer skipping alongside temporal perturbation, as in STG, consistently produces improved results.

\section{Qualitative Comparison}
\label{sec:supple_qual}
We provide additional qualitative comparisons using STG for SVD, Open-Sora, and Mochi. The results demonstrate that applying STG enhances the aesthetic appeal and fidelity of the videos, as shown in Fig.~\ref{fig:opensora_fidelity}.
In Open-Sora, we observe flickering artifacts frequently in the videos. By applying STG, these flickering artifacts are noticeably reduced, as illustrated in Fig.~\ref{fig:temporal_flickering_stg}.

For I2V models such as SVD, as discussed in the main paper, STG not only enhances the structural quality of the generated videos but also increases their dynamic degree. This is because STG mitigates the effect of CFG, which tends to force generated videos to rigidly adhere to the conditioning image. This effect is visualized in Fig.~\ref{fig:dynamic_degree_i2v}.

We provide more video results in the zip file.

\begin{figure}[htbp]
    \centering
    \begin{subfigure}{\linewidth}
        \centering
       {\small \textit{(Image condition is given for SVD.)}}
        \includegraphics[width=\linewidth]{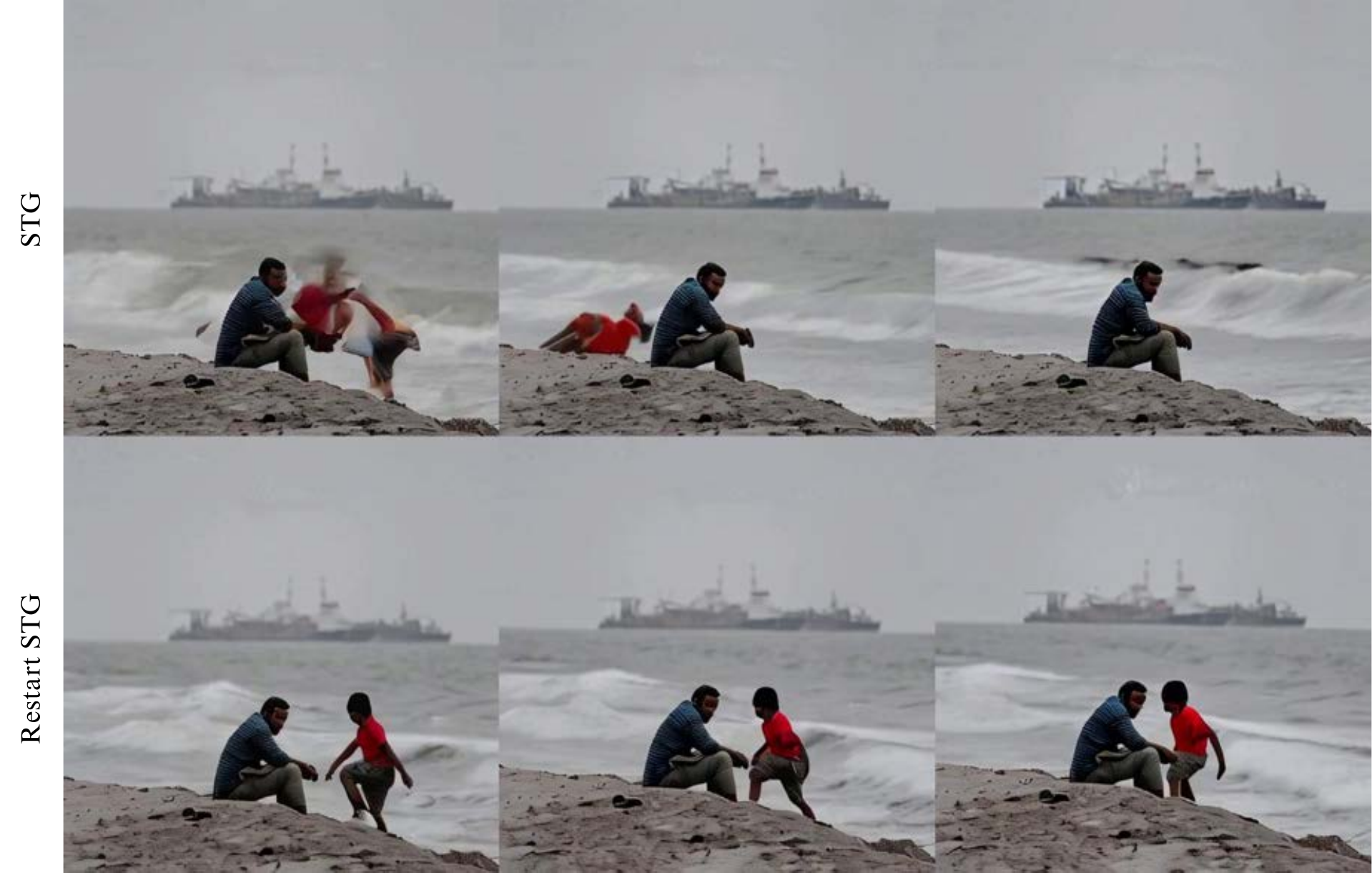}
    \end{subfigure}

    \vspace{0.5cm} 

    \begin{subfigure}{\linewidth}
        \centering
        {\small \textit{Prompt: A group of people sitting on a green bench under an orange tree.}}
        \includegraphics[width=\linewidth]{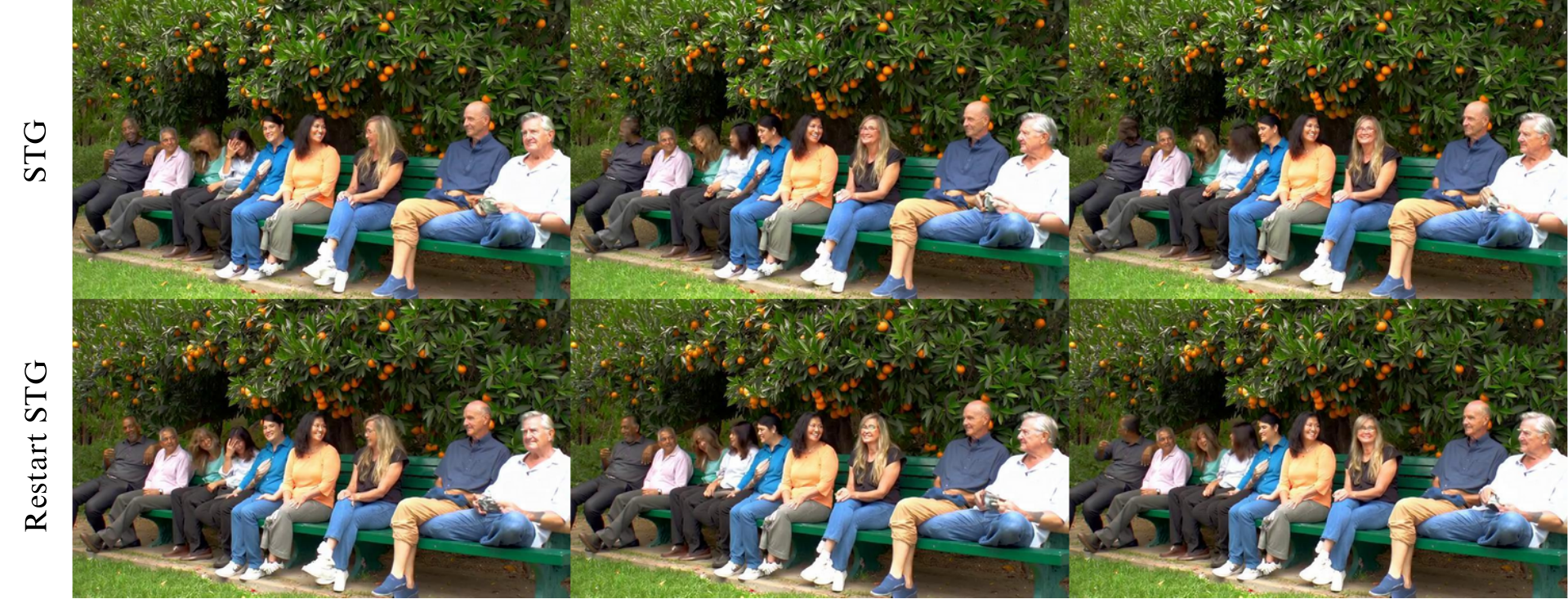}
    \end{subfigure}

    \caption{Quality Improvement with Restart STG. 
    \textit{Top}: Results for SVD~\cite{blattmann2023stable}. 
    \textit{Bottom}: Results for Mochi~\cite{genmo2024mochi}. 
    The results demonstrate that while STG occasionally fails to generate videos correctly in certain cases, applying Restart resolves these issues, producing high-quality and accurate outputs.
}
    \label{fig:restart_stg}
\end{figure}

\begin{figure*}
    \centering
    {\small \textit{Prompt: A young woman with glasses is jogging in the park wearing a pink headband.}} 
    \includegraphics[width=\linewidth]{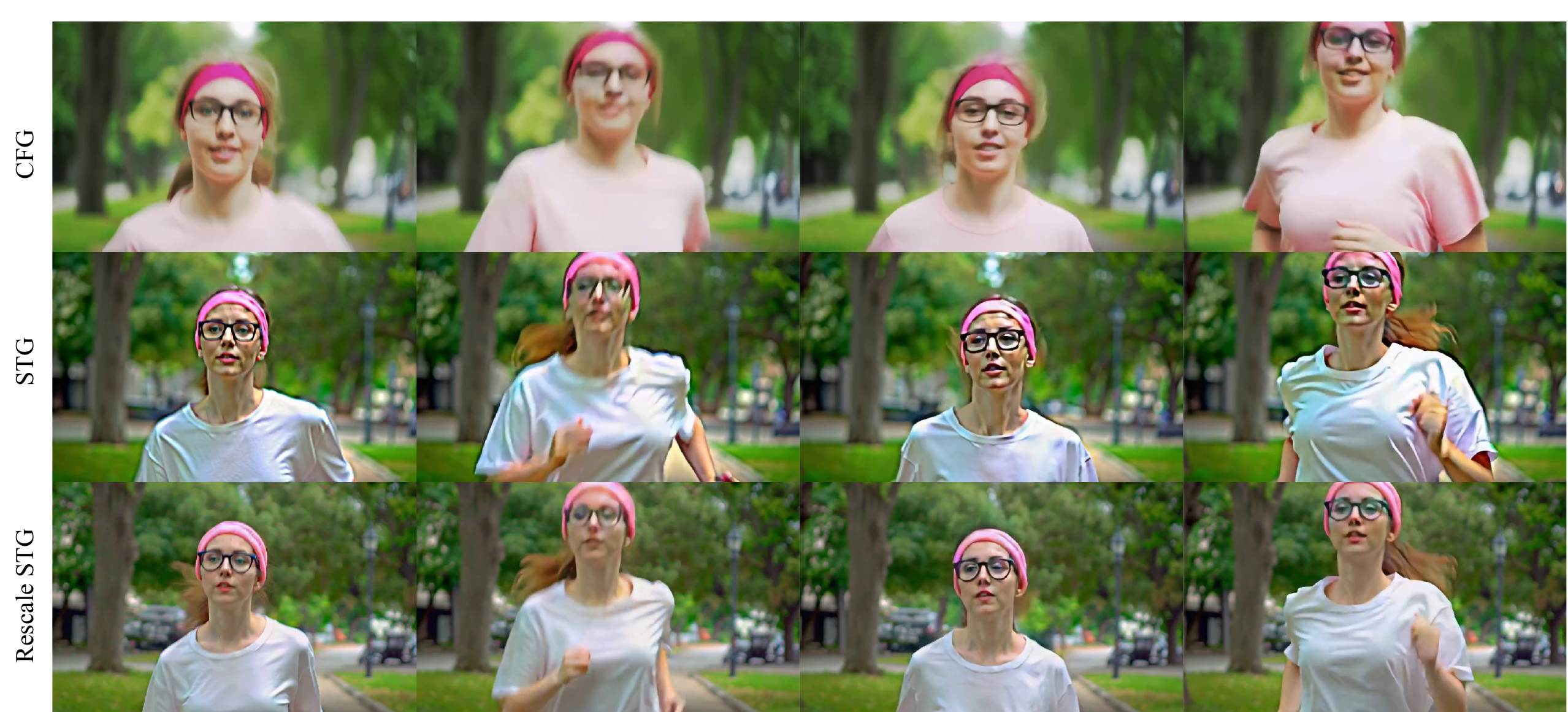}
    \caption{Comparison of CFG, STG, and Rescaled STG on Mochi~\cite{genmo2024mochi}. When STG is applied using large scales or to bright videos, it often suffers from over-saturation caused by manifold deviation. One potential cause of this issue is the increased variance in the latent code, which is effectively mitigated by the rescaling technique proposed in ~\cite{lin2024common}.
    }
    \label{fig:rescale_stg}
    \vspace{-2mm}
\end{figure*}

\begin{figure*}
    \centering
    {\small \textit{Prompt: A close-up shot of a butterfly landing on the nose of a woman, highlighting her smile and the details of the butterfly's wings.}} 
    \includegraphics[width=\linewidth]{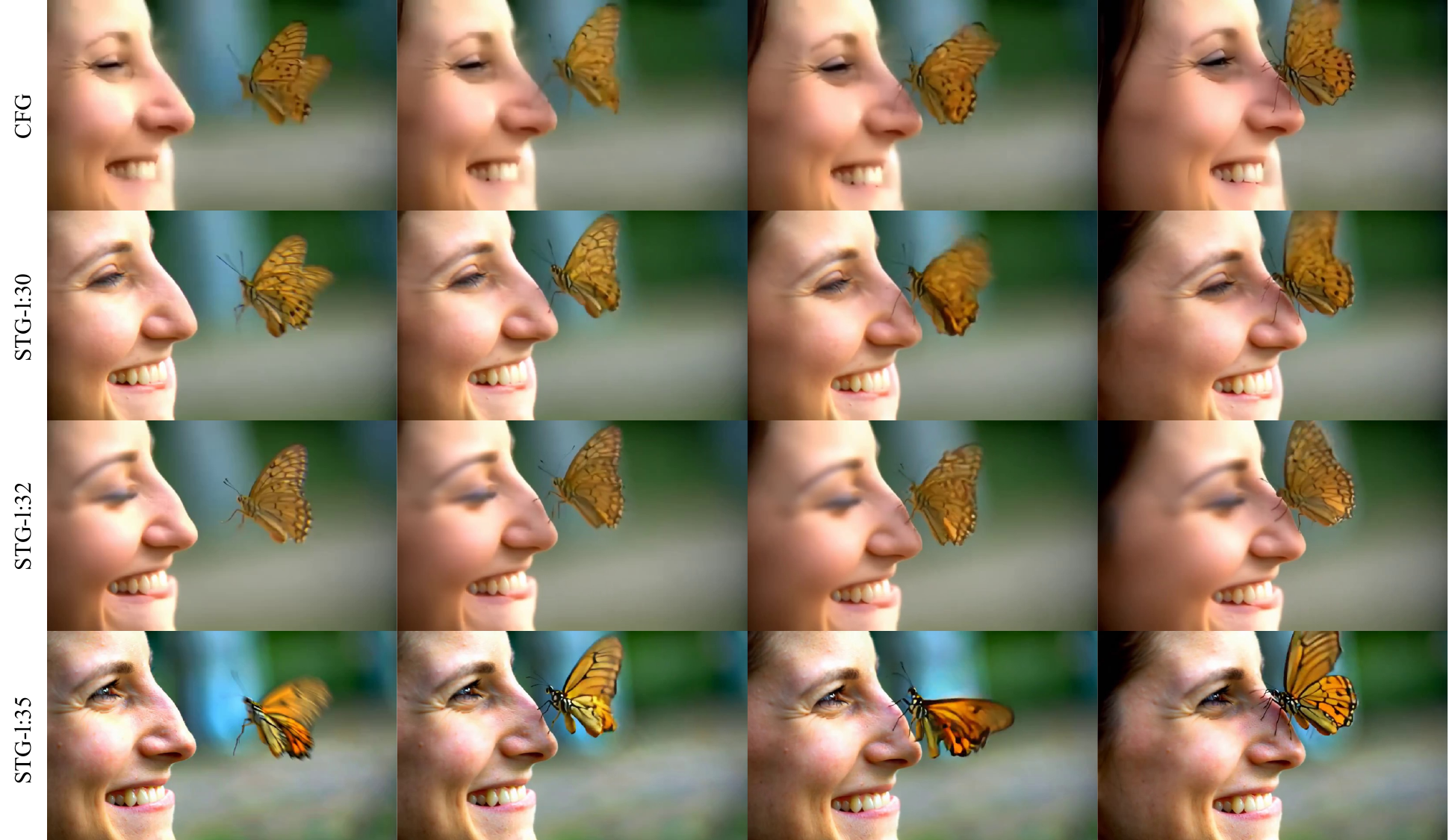}
    \caption{Ablation study on the effect of skipping different layers for STG on Mochi~\cite{genmo2024mochi}. Our results indicate that skipping later layers has a greater impact on quality improvements, as these layers primarily contribute to texture details. For all experiments, we consistently skip layer 35 (denoted as STG-l:35).
    }
    \label{fig:layer-ab}
    \vspace{-4mm}
\end{figure*}

\begin{figure}[htbp]
    \centering
    {\small \textit{(Image condition is given for SVD.)}}
    \includegraphics[width=\linewidth]{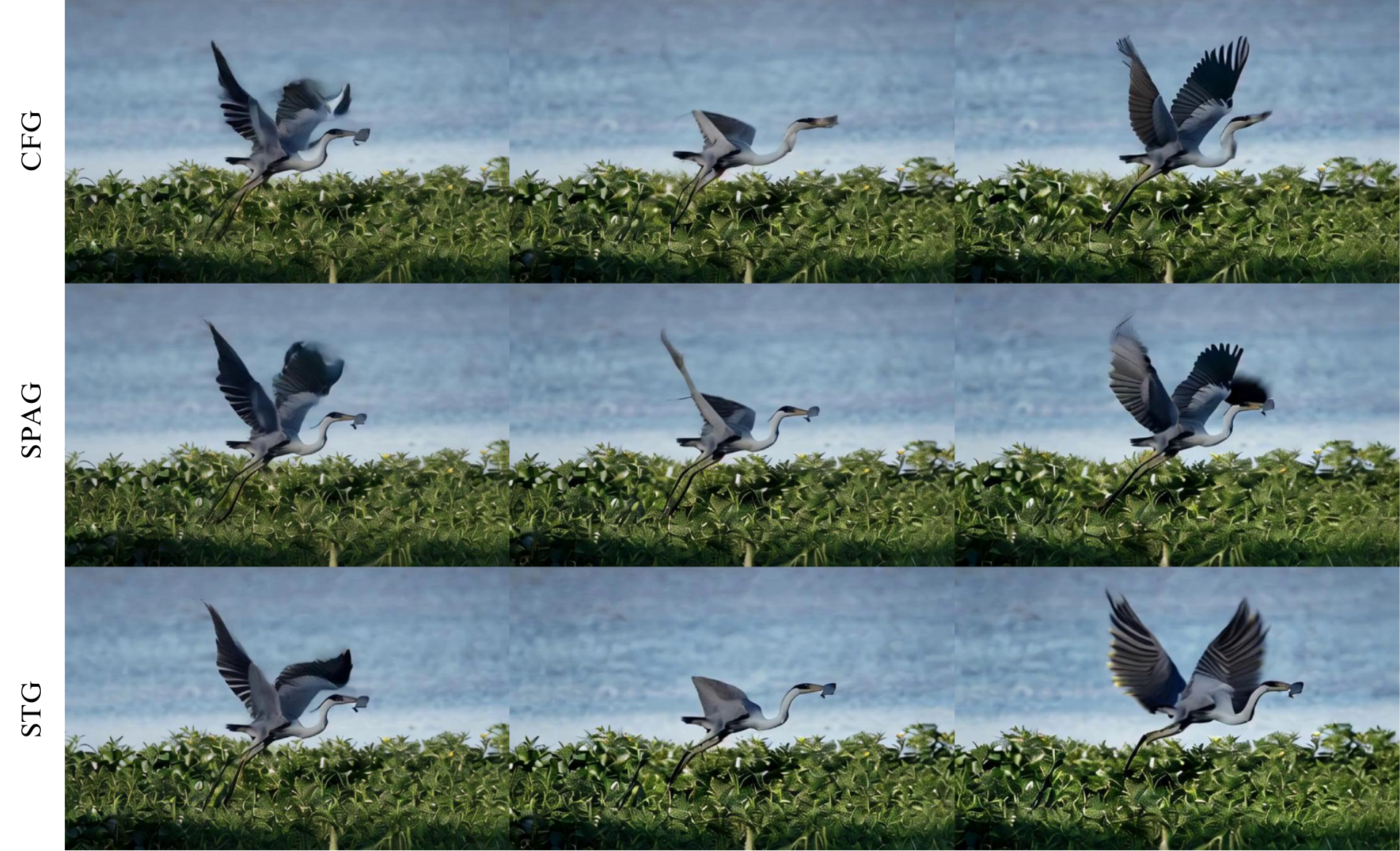}
    \caption{Qualitative Comparison of CFG, SPAG, and STG on SVD~\cite{blattmann2023stable}. PAG applied only to spatial layers is referred to as SPAG. The results show that while CFG and SPAG fail to preserve object clarity under motion, STG successfully achieves this.}
    \label{fig:cfg_pag_stg_svd}
\end{figure}

\begin{figure}[htbp]
    \centering
    {\small \textit{Prompt: Brown chicken hunting for its food.}}
    \includegraphics[width=\linewidth]{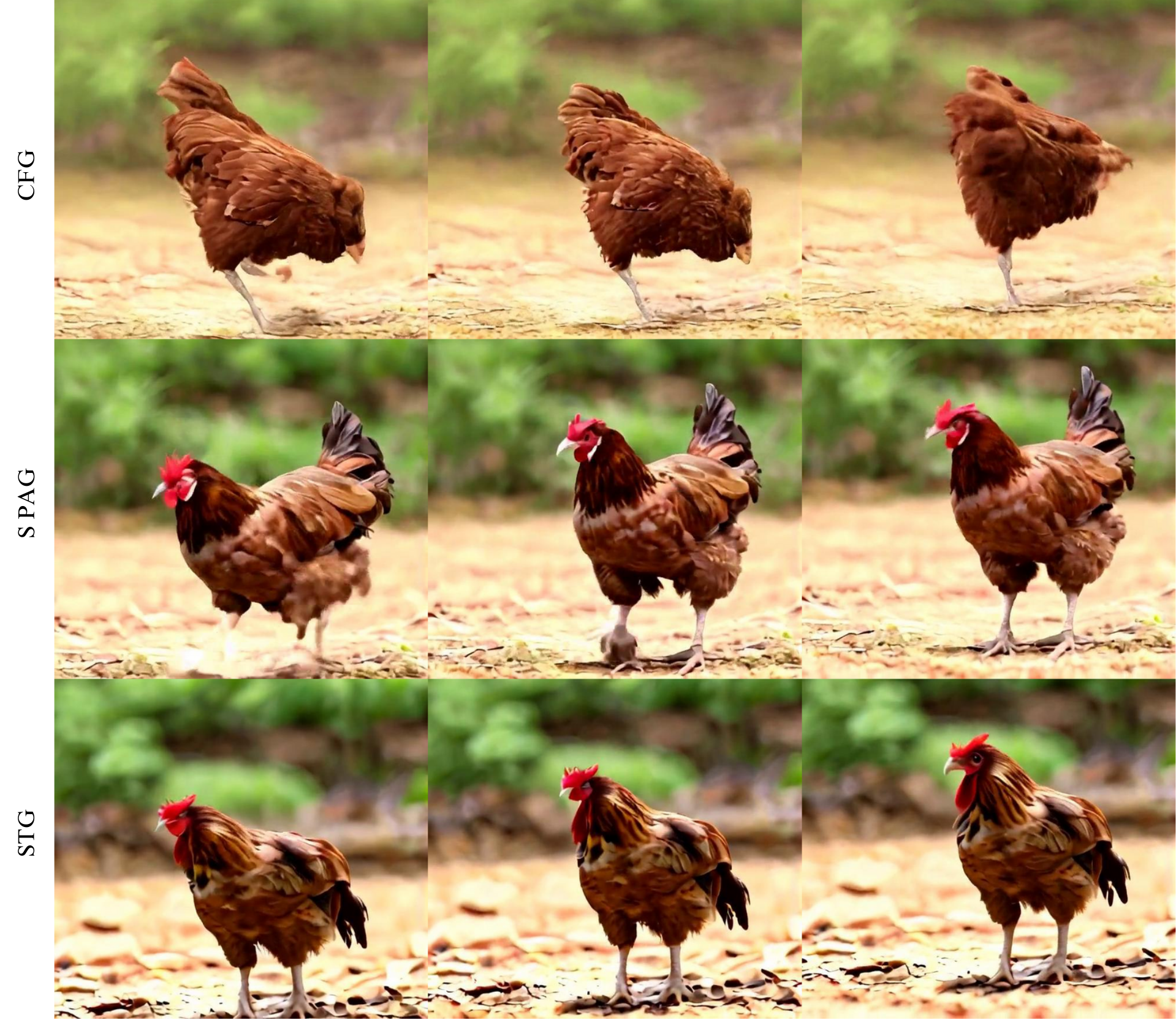}
    \caption{
Qualitative Comparison of CFG, SPAG, and STG on Open-Sora~\cite{opensora}. The results show CFG fails to generate the object's head accurately, and SPAG struggles with the legs, whereas STG successfully generates all components correctly.}
    \label{fig:cfg_pag_stg_opensora}
\end{figure}

\begin{figure}[htbp]
    \centering
    {\small \textit{(Image condition is given for SVD.)}}
    \includegraphics[width=\linewidth]{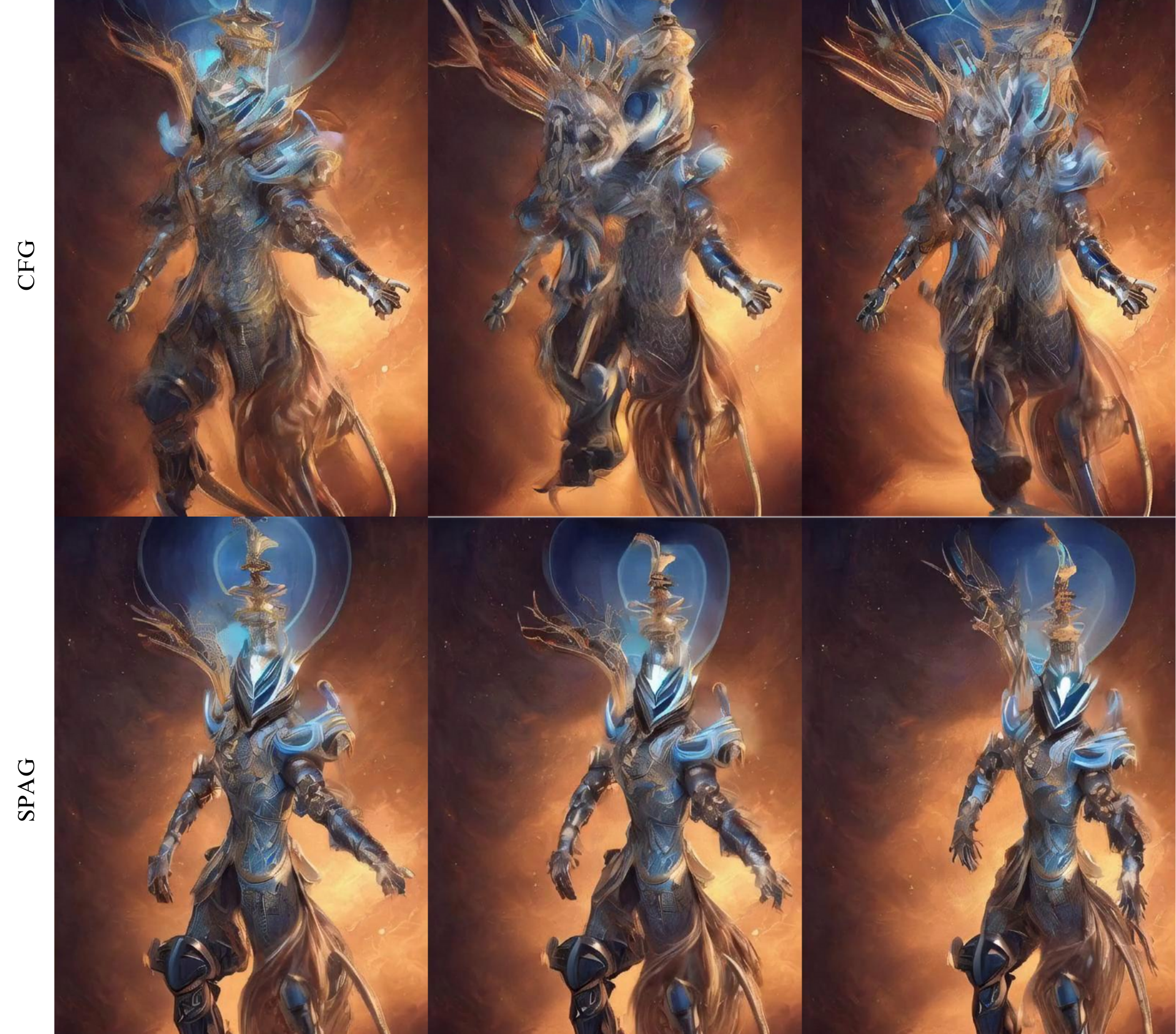}
    \caption{Qualitative Comparison of Object Structure in SVD~\cite{blattmann2023stable} with and without Spatial Guidance. Spatial Guidance is represented by SPAG, which applies PAG only to the spatial layer. 
    The results indicate that while CFG struggles to maintain clear object structures, leading to blurry videos, SPAG effectively enhances object structure and improves clarity.}
    \label{fig:spag}
\end{figure}

\begin{figure}[htbp]
    \centering
    
\end{figure}
\begin{figure}[htbp]
    \centering
    {\small \textit{(Image condition is given for SVD.)}}
    \includegraphics[width=\linewidth]{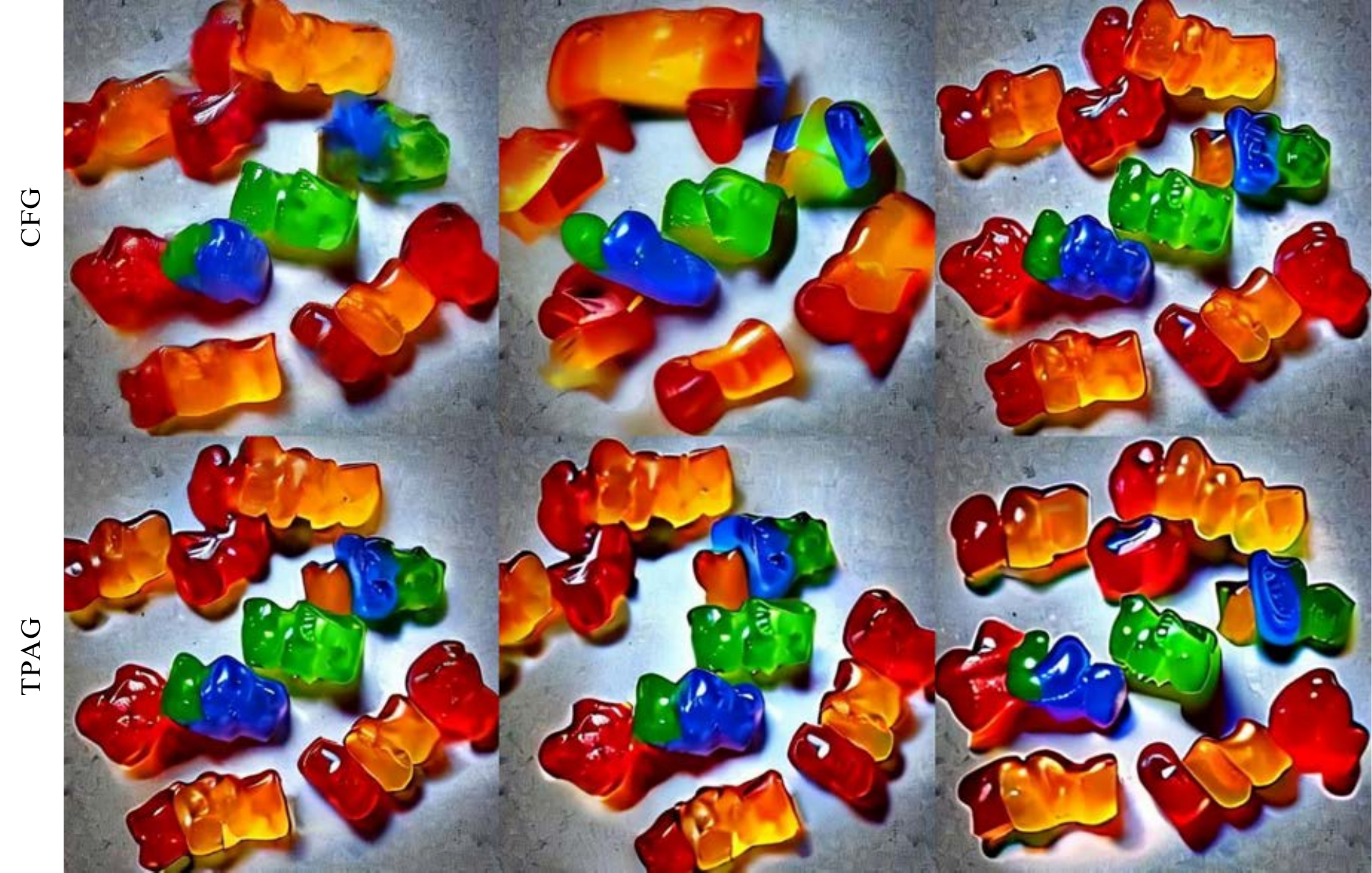}
    \caption{Qualitative Comparison of Temporal Consistency in SVD~\cite{blattmann2023stable} with and without Temporal Guidance (TPAG). The results reveal that CFG struggles to ensure frame-to-frame consistency, with the shape and color of the jelly varying noticeably across frames, leading to a disjointed video. In contrast, TPAG effectively preserves the jelly's appearance throughout the sequence, creating a more cohesive video and significantly improving Temporal Consistency.
    }
    \label{fig:tpag}

\end{figure}

\begin{figure}[htbp]
    \centering
    \begin{subfigure}[b]{\linewidth}
        \centering
        {\small \textit{Prompt: A close-up shot of a butterfly landing on the nose of a woman, highlighting her smile and the details of the butterfly's wings.}}
        \includegraphics[width=\linewidth]{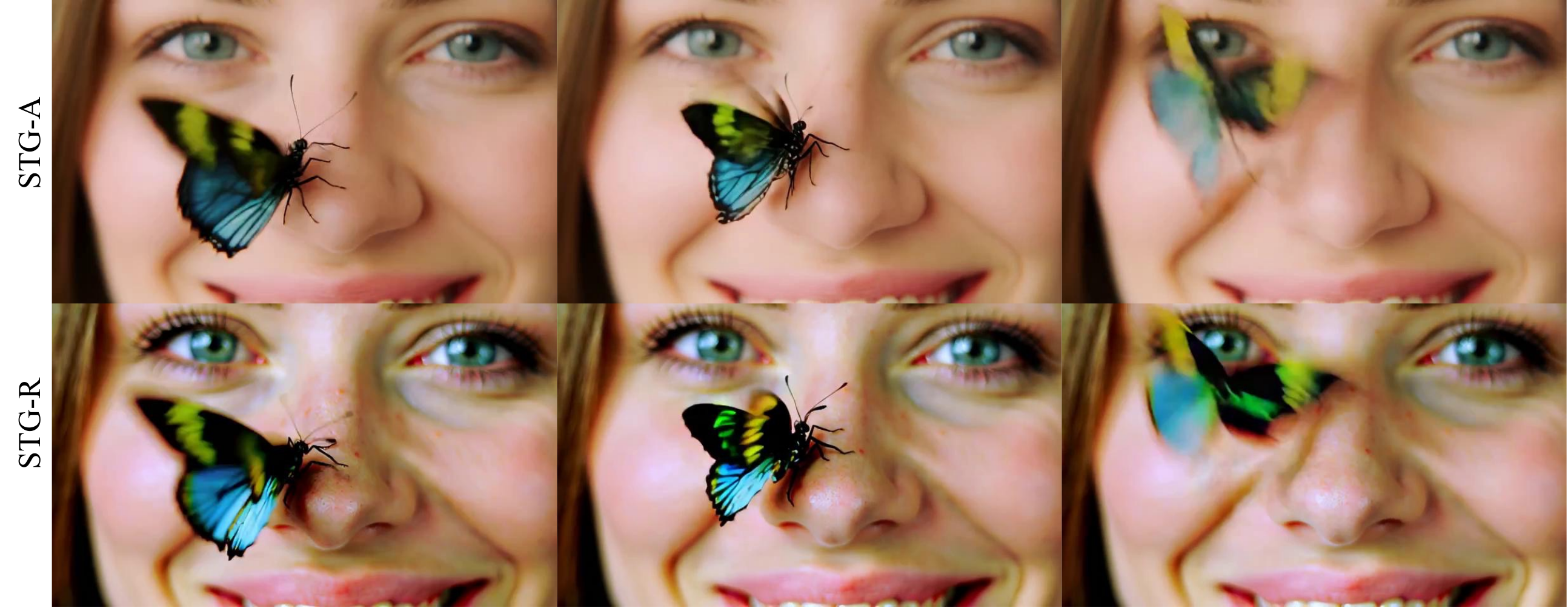}
    \end{subfigure}
    
    \vspace{0.5cm}
    
    \begin{subfigure}[b]{\linewidth}
        \centering
        {\small \textit{Prompt: Cinematic 8k scene of a couple dancing under warmly glowing string lights in an intimate backyard setting, ...}}
        \includegraphics[width=\linewidth]{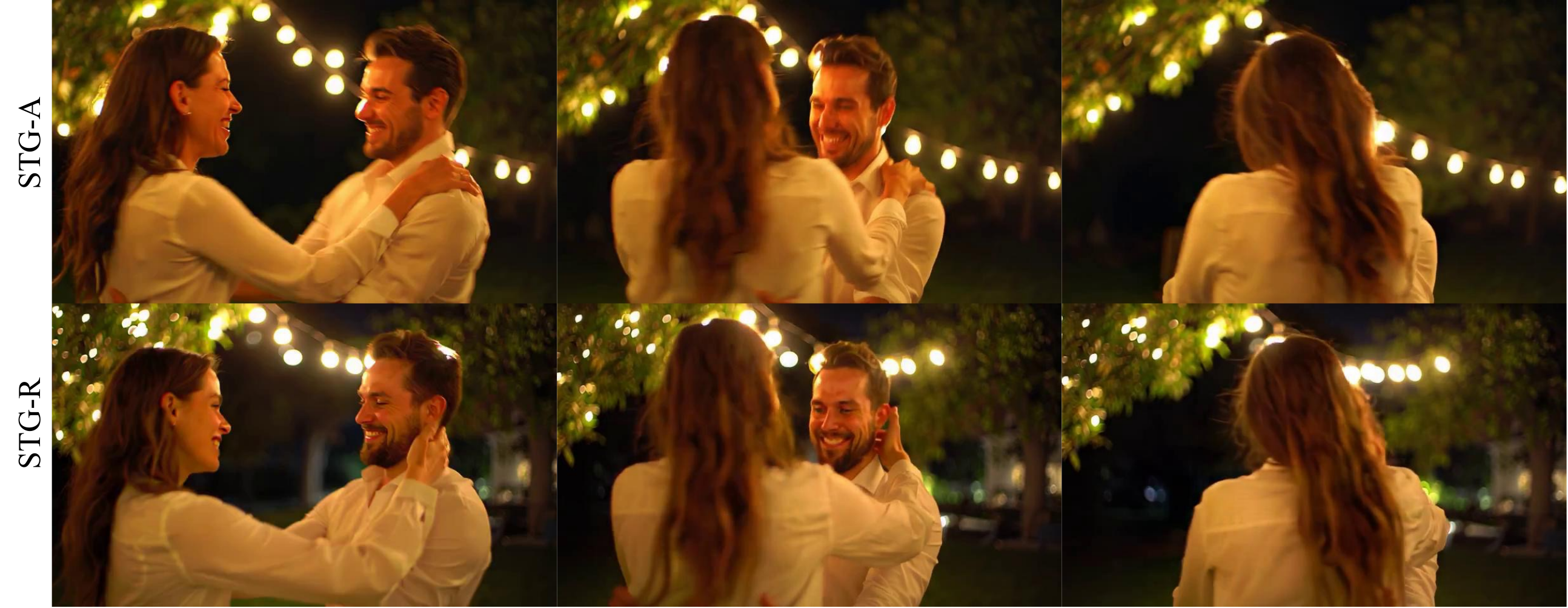}
    \end{subfigure}
    \caption{Comparison of attention skip (STG-A) and residual skip (STG-R) in Mochi~\cite{genmo2024mochi}. The results indicate that STG-R delivers greater qualitative improvements for Mochi.}
    \label{fig:stgar_mochi}
\end{figure}

\begin{figure}[htbp]
    \centering
    \begin{subfigure}[b]{\linewidth}
        \centering
        {\small \textit{Prompt: A close-up portrait of a woman set against a snowy backdrop. The woman is wearing a golden crown...}}
        \includegraphics[width=\linewidth]{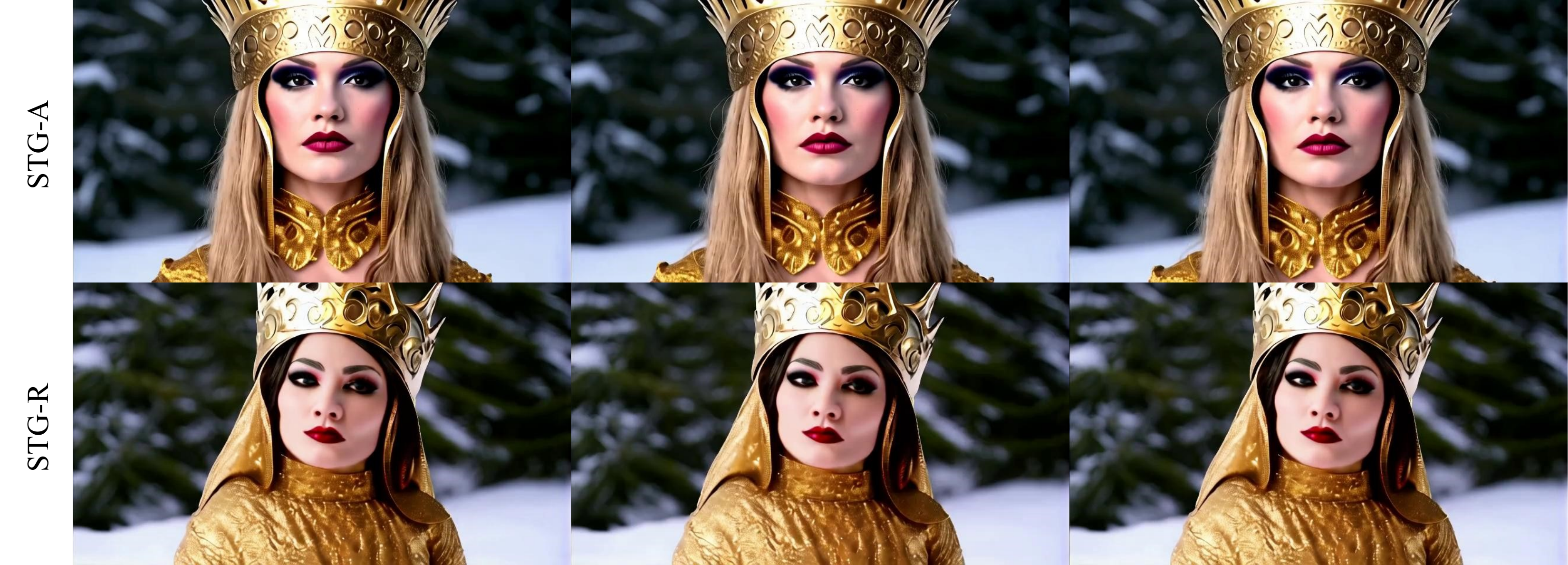}
    \end{subfigure}
    
    \vspace{0.5cm}
    
    \begin{subfigure}[b]{\linewidth}
        \centering
        {\small \textit{Prompt: A moment of a woman in a white wedding dress, adorned with a pearl necklace and veil, standing...}}
        \includegraphics[width=\linewidth]{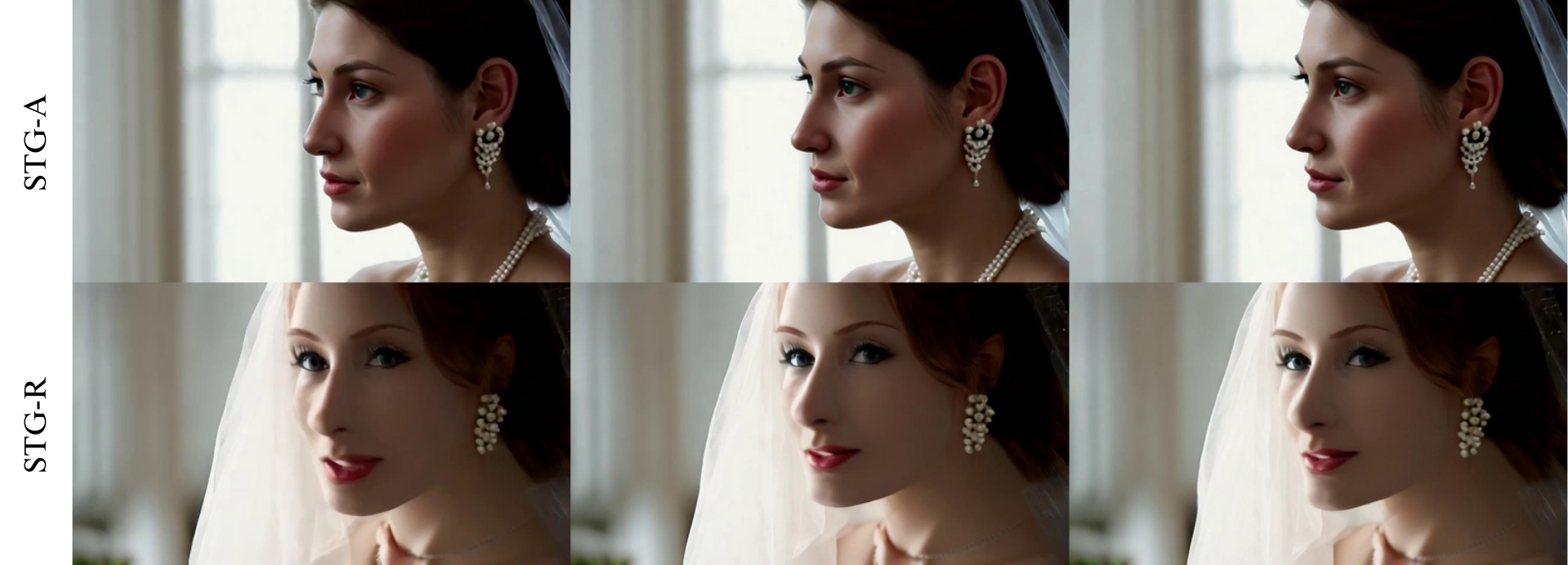}
    \end{subfigure}
    \caption{Comparison of attention skip (STG-A) and residual skip (STG-R) in Open-Sora~\cite{opensora}. The results indicate that STG-A delivers greater qualitative improvements for Open-Sora.}
    \label{fig:stgar_opensora}
\end{figure}

\begin{figure}[htbp]
    \centering
    \begin{subfigure}[b]{\linewidth}
        \centering
        {\small \textit{(Image condition is given for SVD.)}}
        \includegraphics[width=\linewidth]{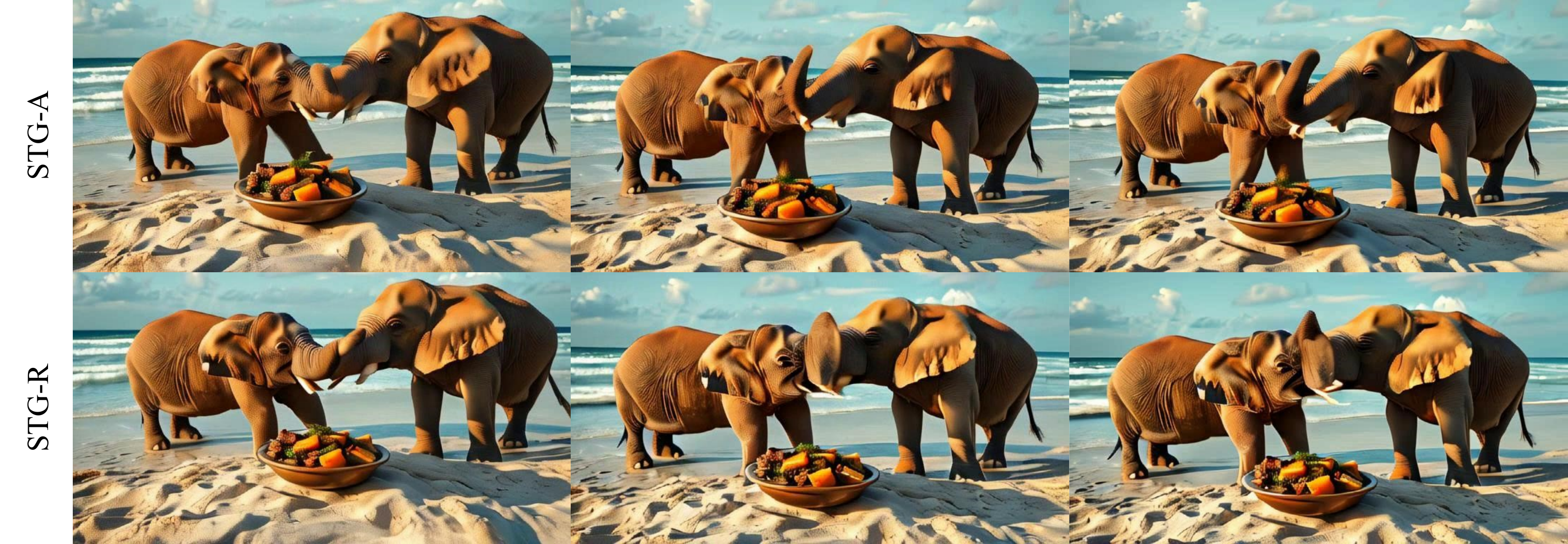}
    \end{subfigure}
    
    \vspace{0.5cm}
    
    \begin{subfigure}[b]{\linewidth}
        \centering
        {\small \textit{(Image condition is given for SVD.)}}
        \includegraphics[width=\linewidth]{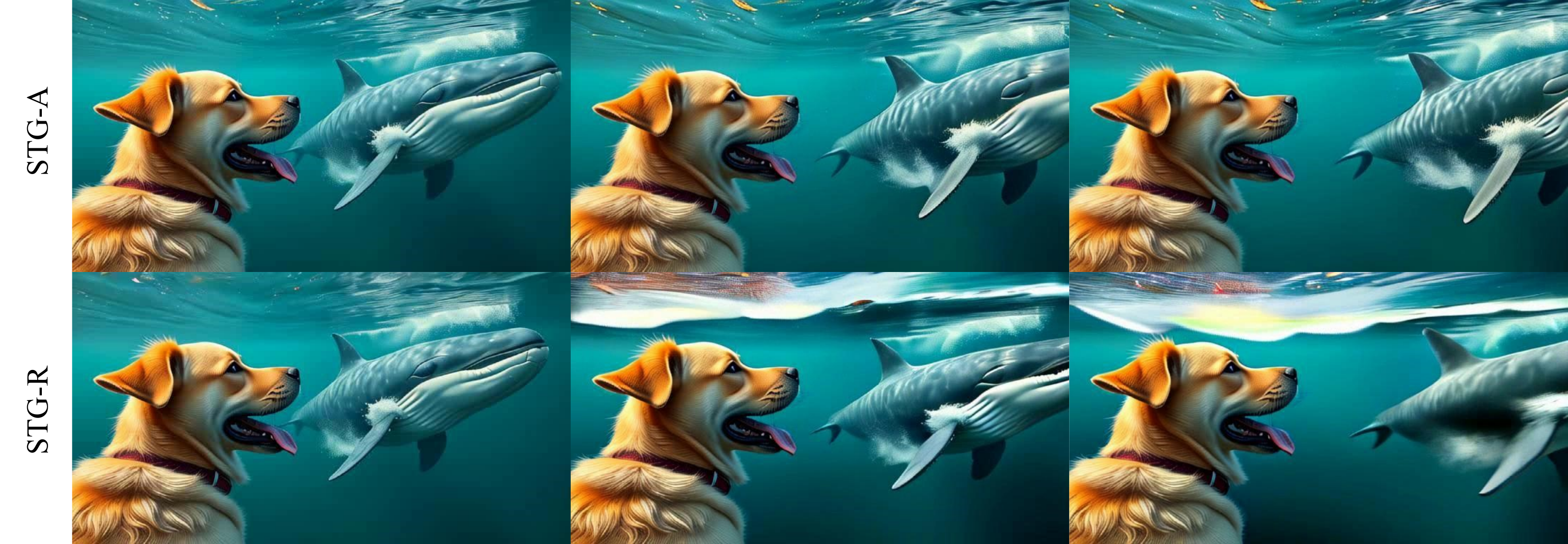}
    \end{subfigure}
    \caption{Comparison of attention skip (STG-A) and residual skip (STG-R) in SVD~\cite{blattmann2023stable}. The results indicate that STG-A delivers greater qualitative improvements for SVD.}
    \label{fig:stgar_svd}
\end{figure}

\begin{figure}[htbp]
    \centering
    {\small \textit{(Image condition is given for SVD.)}}
    \includegraphics[width=\linewidth]{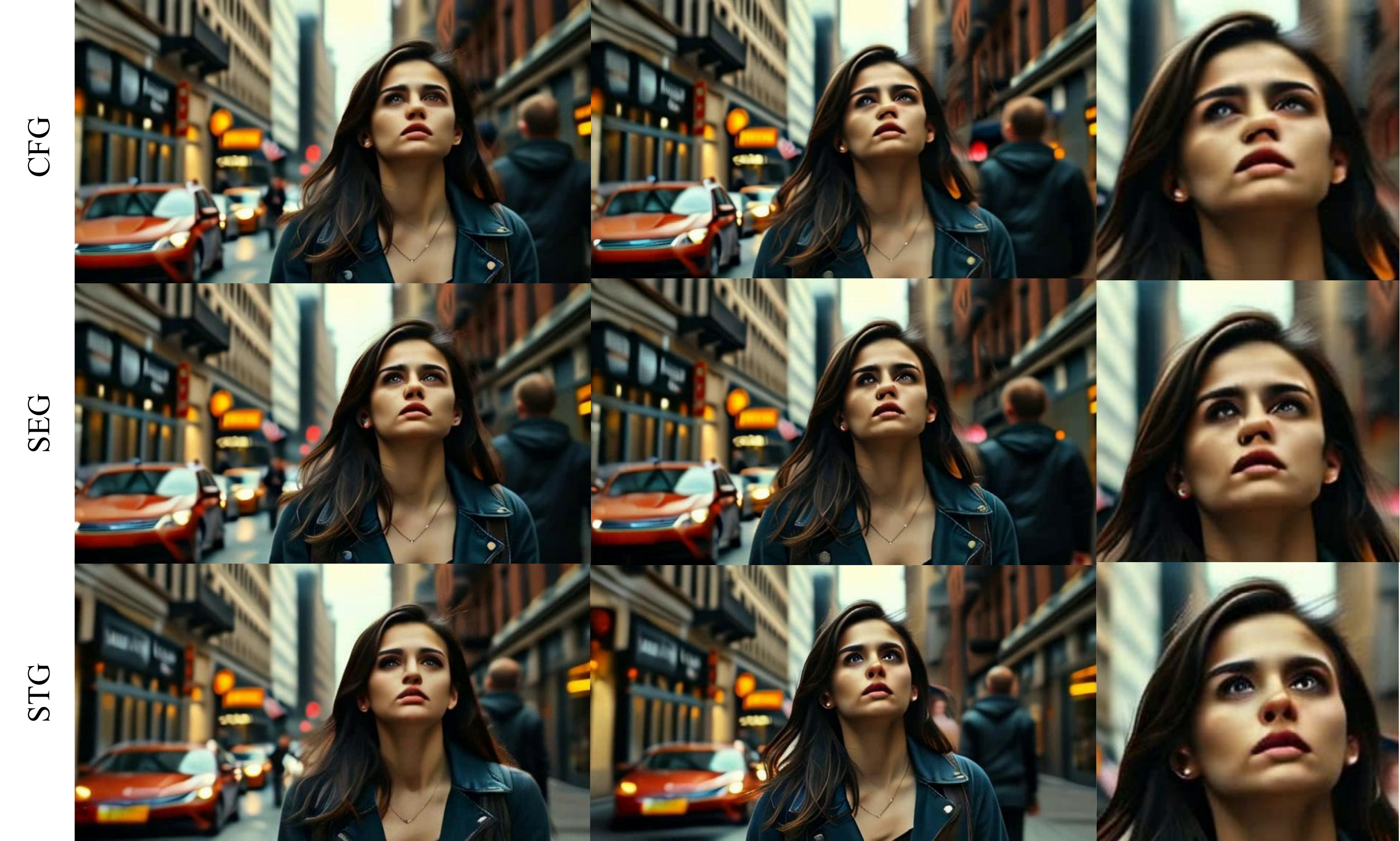}
    \caption{Comparison of CFG, SEG~\cite{hong2024smoothed}, and STG in SVD~\cite{blattmann2023stable}. The results show that CFG and SEG generate an unnatural nose for the person, whereas STG successfully generates all components naturally.}
    \label{fig:seg_comp}
\end{figure}

\begin{figure*}
    \centering
    {\small \textit{Prompt: A family having a picnic under a shady tree in a large park.}}
    \includegraphics[width=0.9\linewidth]{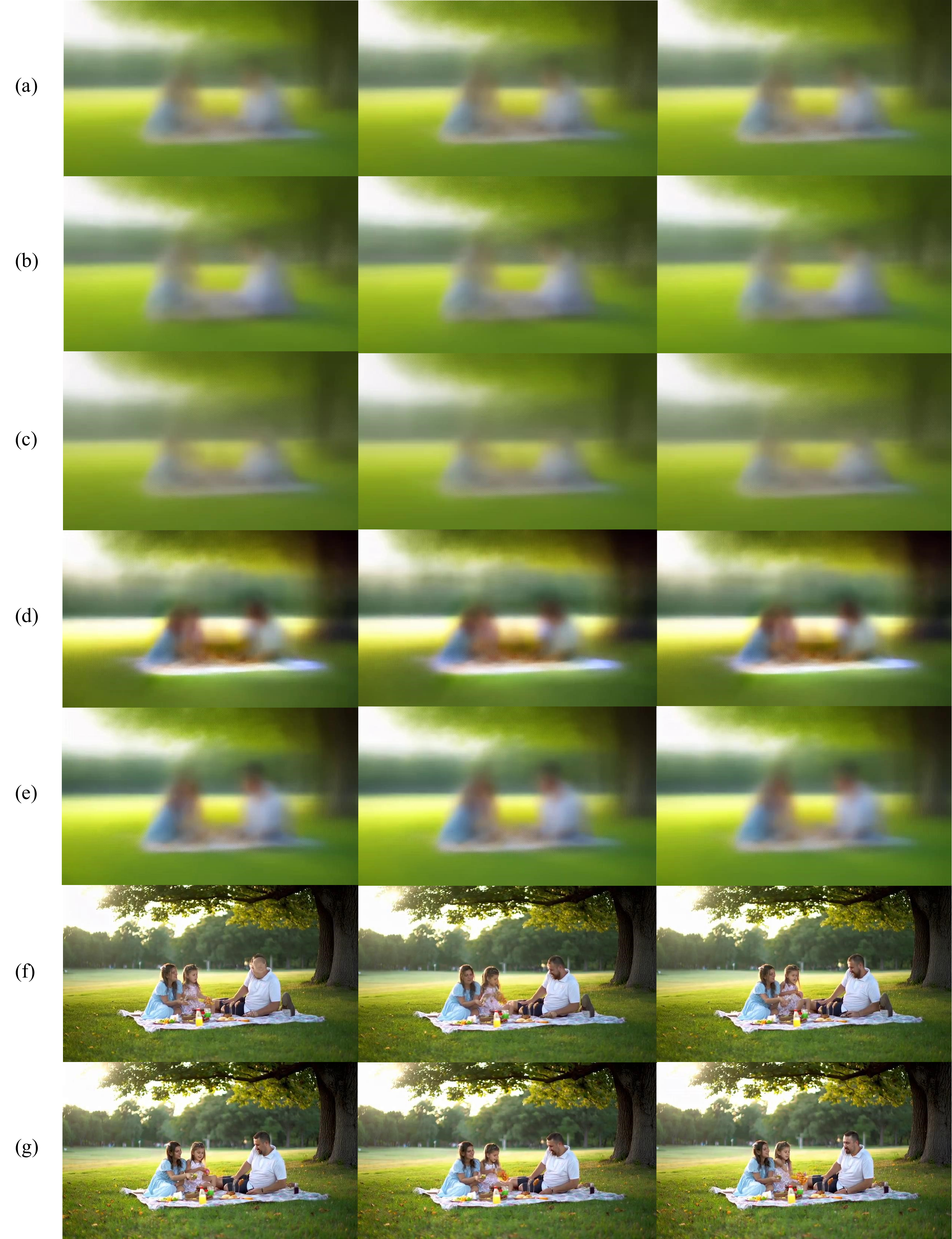}
    \caption{Weak model visualization for Mochi~\cite{genmo2024mochi}. Generated video using one-step prediction from timestep 30 (\(t=30\)). 
        (a) \(\epsilon_\theta(x_t)\), (b) \(\epsilon_\theta(x_t | \phi)\), (c) \(\epsilon_\theta^{s, t}(x_t)\), (d) CFG, (e) STG (f) Final video (CFG) (g) Final video (STG). 
        The video predicted by CFG exhibits unnatural colors in certain areas and broken structures. In contrast, the video generated with STG demonstrates improved structural integrity and more natural color tones.}
    \label{fig:weak_t30}
    \vspace{-4mm}
\end{figure*}

\begin{figure*}
    \centering
    {\small \textit{Prompt: A family having a picnic under a shady tree in a large park.}}
    \includegraphics[width=0.9\linewidth]{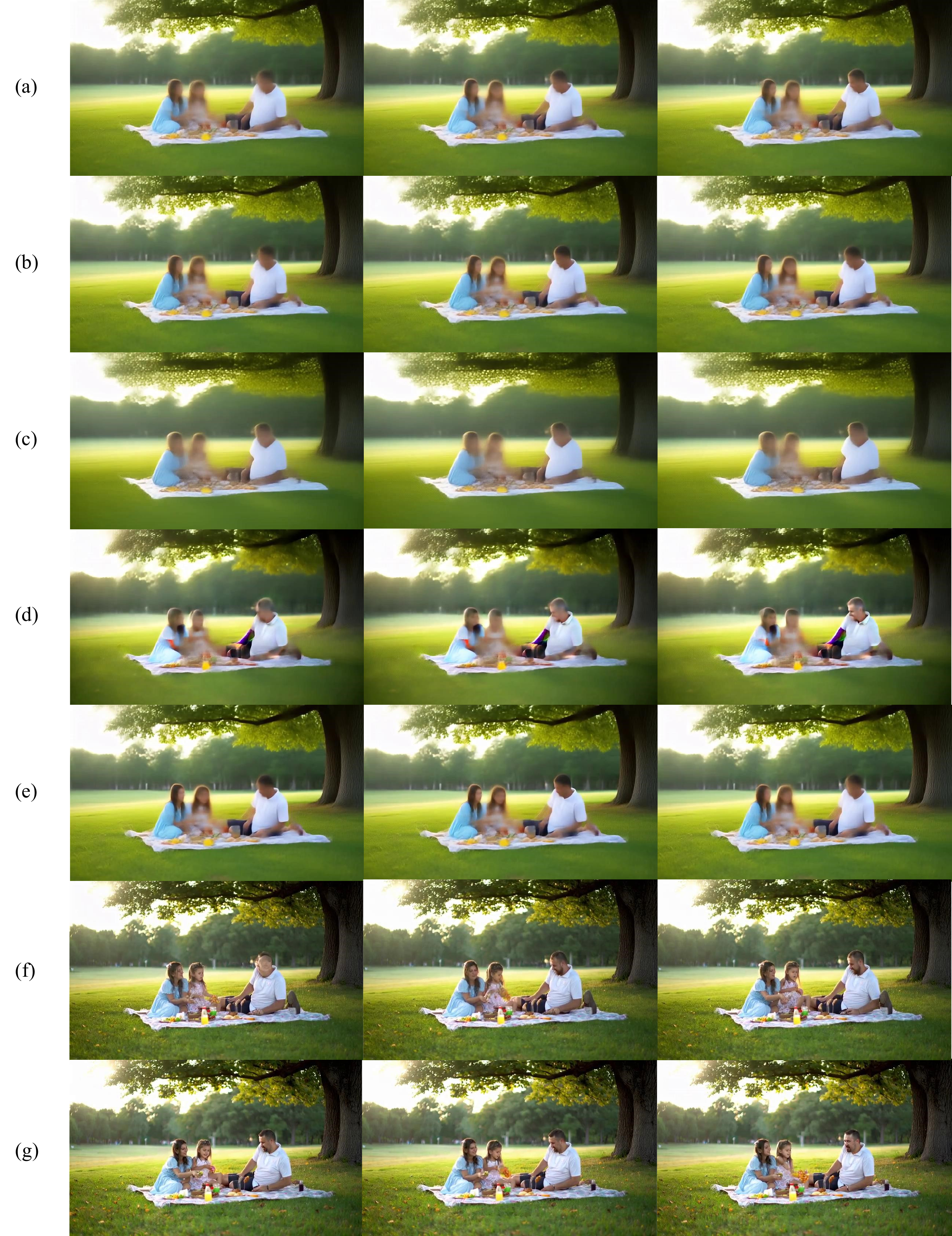}
    \caption{Weak model visualization for Mochi~\cite{genmo2024mochi}. Video generated using one-step prediction from timestep 24 (\(t=24\)). 
    (a) \(\epsilon_\theta(x_t)\), (b) \(\epsilon_\theta(x_t | \phi)\), (c) \(\epsilon_\theta^{s, t}(x_t)\), (d) CFG, (e) STG (f) Final video (CFG) (g) Final video (STG). 
    The result demonstrates that STG effectively guides the model to maintain structural integrity and realistic color distribution while avoiding the unintended artifacts present in CFG predictions.
    }
    \label{fig:weak_t24}
    \vspace{-4mm}
\end{figure*}

\begin{figure}[htbp]
\centering
\begin{subfigure}{\linewidth}
\centering
{\small \textit{Prompt: a neon-lit cityscape at night, featuring towering skyscrapers and crowded streets. The streets are bustling...}}
\includegraphics[width=\linewidth]{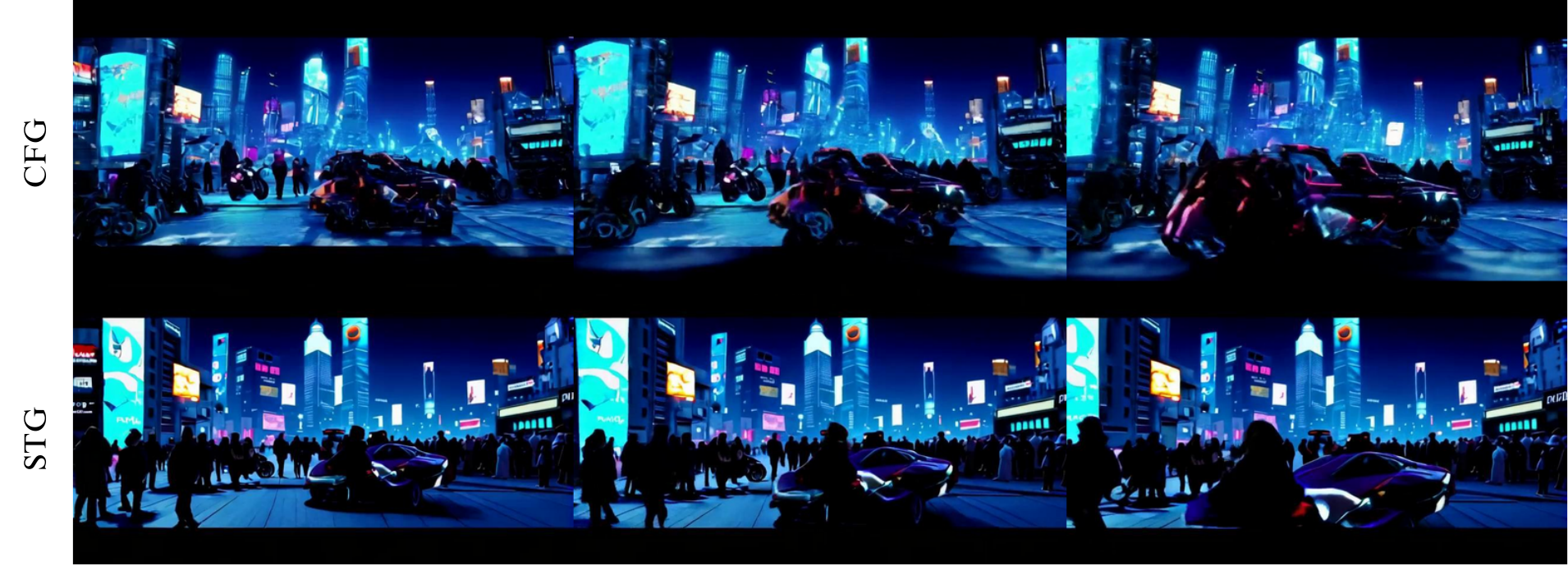}
\end{subfigure}

\vspace{0.5cm} 

\begin{subfigure}{\linewidth}
\centering
{\small \textit{Prompt: A fluffy grey and white cat is lazily stretched out on a sunny window sill, enjoying a nap after a long day of lounging.}}
\includegraphics[width=\linewidth]{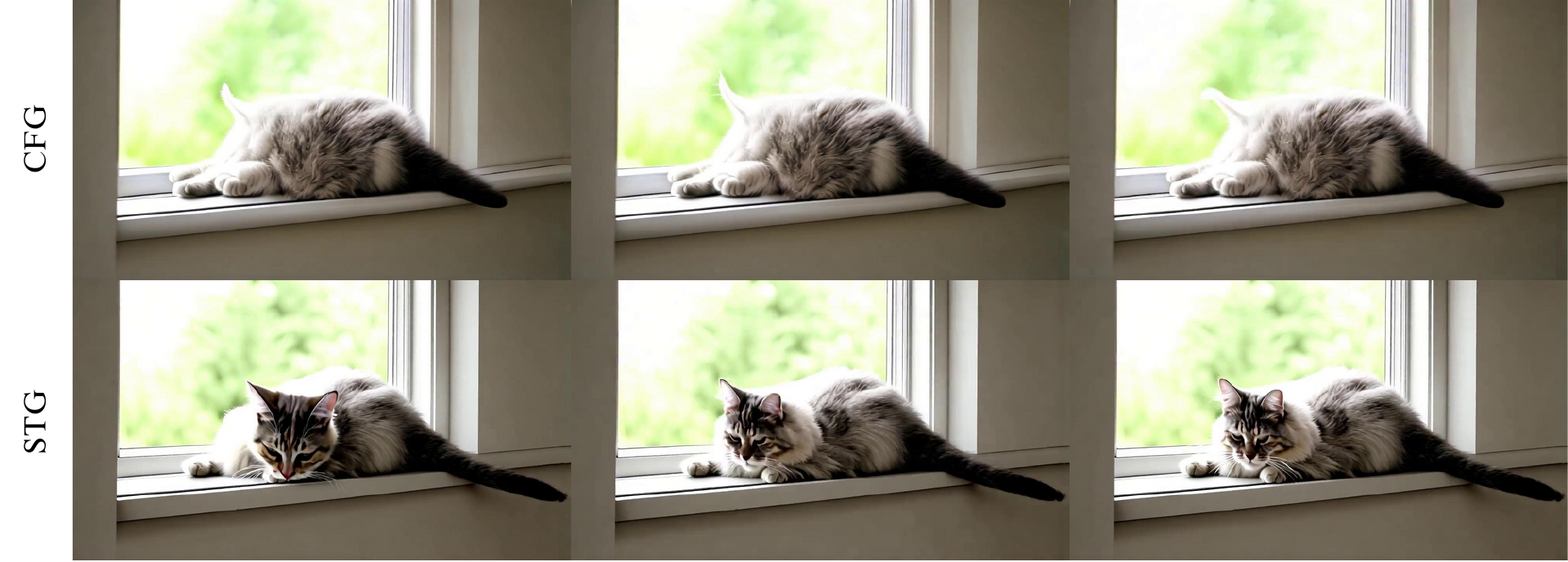}
\end{subfigure}

\vspace{0.5cm} 

\begin{subfigure}{\linewidth}
\centering
{\small \textit{Prompt: Iron Man is walking towards the camera in the rain at night, with a lot of fog behind him. Science fiction movie, close-up.}}
\includegraphics[width=\linewidth]{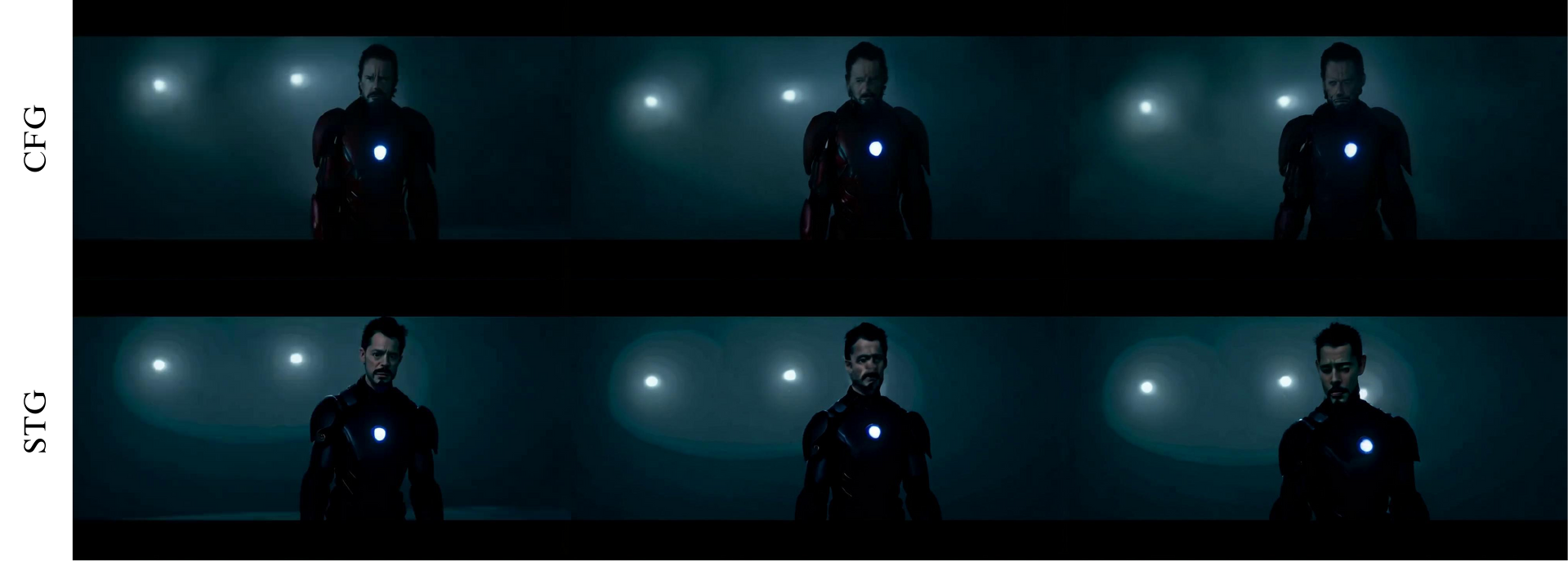}
\end{subfigure}

\caption{Qualitative comparison of video quality with and without STG applied on Open-Sora~\cite{opensora}. The results demonstrate that applying STG enhances the video's aesthetic appeal and fidelity.}
\label{fig:opensora_fidelity}
\end{figure}

\begin{figure}[htbp]
    \centering
    \begin{subfigure}{\linewidth}
        \centering
        {\small \textit{Prompt: A dog wearing vr goggles on a boat.}}
        \includegraphics[width=\linewidth]{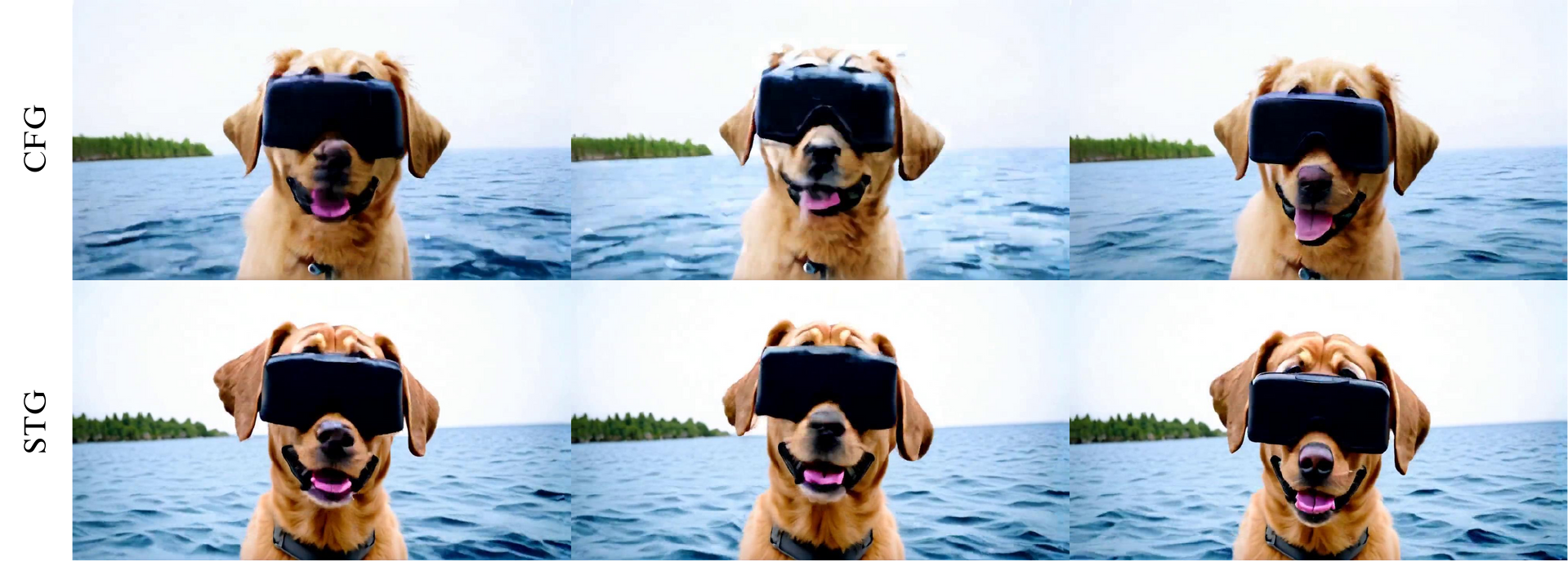}
    \end{subfigure}

    \vspace{0.5cm} 

    \begin{subfigure}{\linewidth}
        \centering
        {\small \textit{Prompt: A cyborg standing on top of a skyscraper, overseeing the city, back view, cyberpunk vibe, 2077, NYC, intricate details, 4K.}}
        \includegraphics[width=\linewidth]{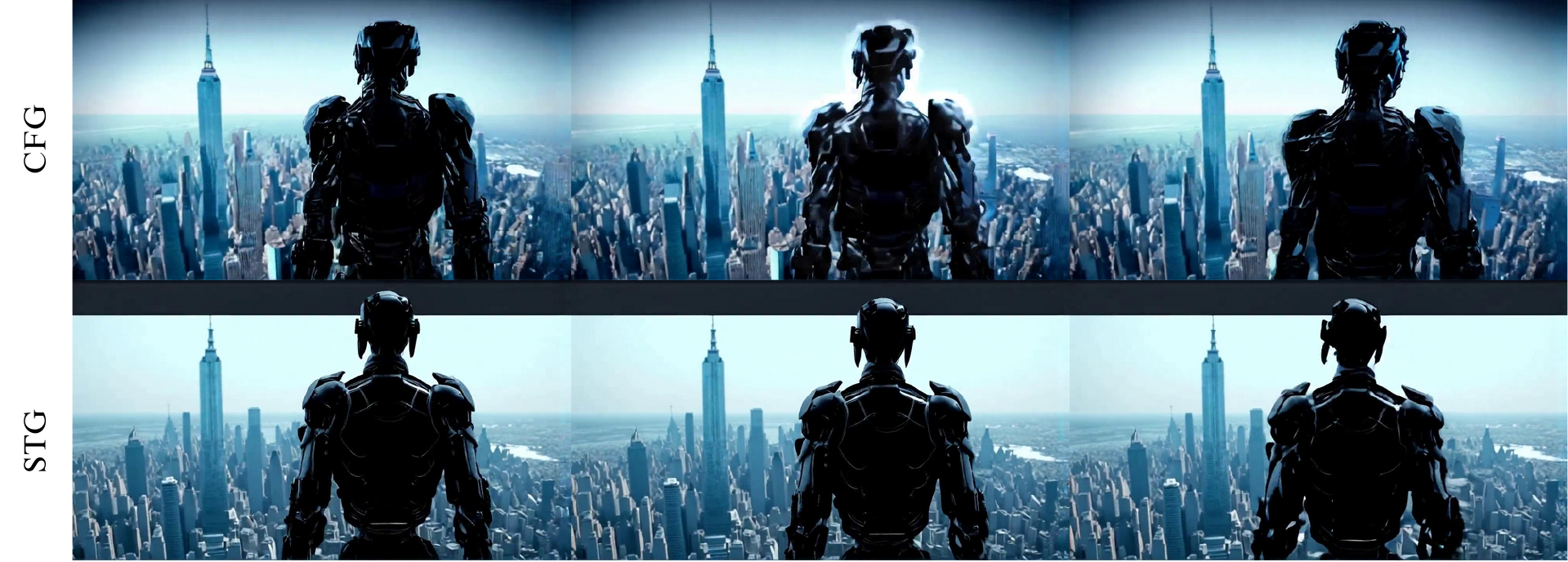}
    \end{subfigure}

    \caption{
Qualitative comparison of Temporal Flickering in Open-Sora~\cite{opensora} with and without STG. Without STG, temporal flickering is observed around the object, causing sudden bright flashes that disrupt the video experience. STG significantly reduces these artifacts, resulting in smoother and more cohesive motion.}
    \label{fig:temporal_flickering_stg}
\end{figure}

\begin{figure}[htbp]
    \centering
    \begin{subfigure}{\linewidth}
        \centering
        {\small \textit{(Image condition is given for SVD.)}}
        \includegraphics[width=\linewidth]{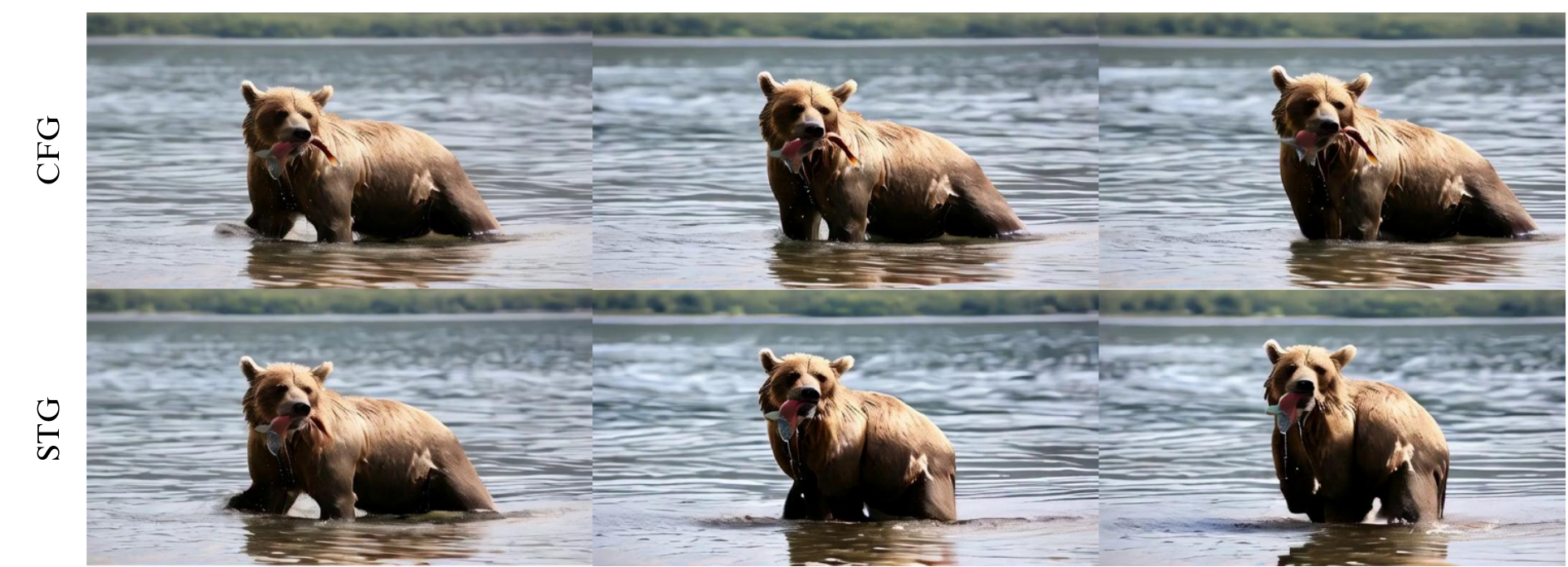}
    \end{subfigure}
    
    \vspace{0.5cm} 

    \begin{subfigure}{0.9\linewidth}
        \centering
        {\small \textit{(Image condition is given for SVD.)}}
        \includegraphics[width=\linewidth]{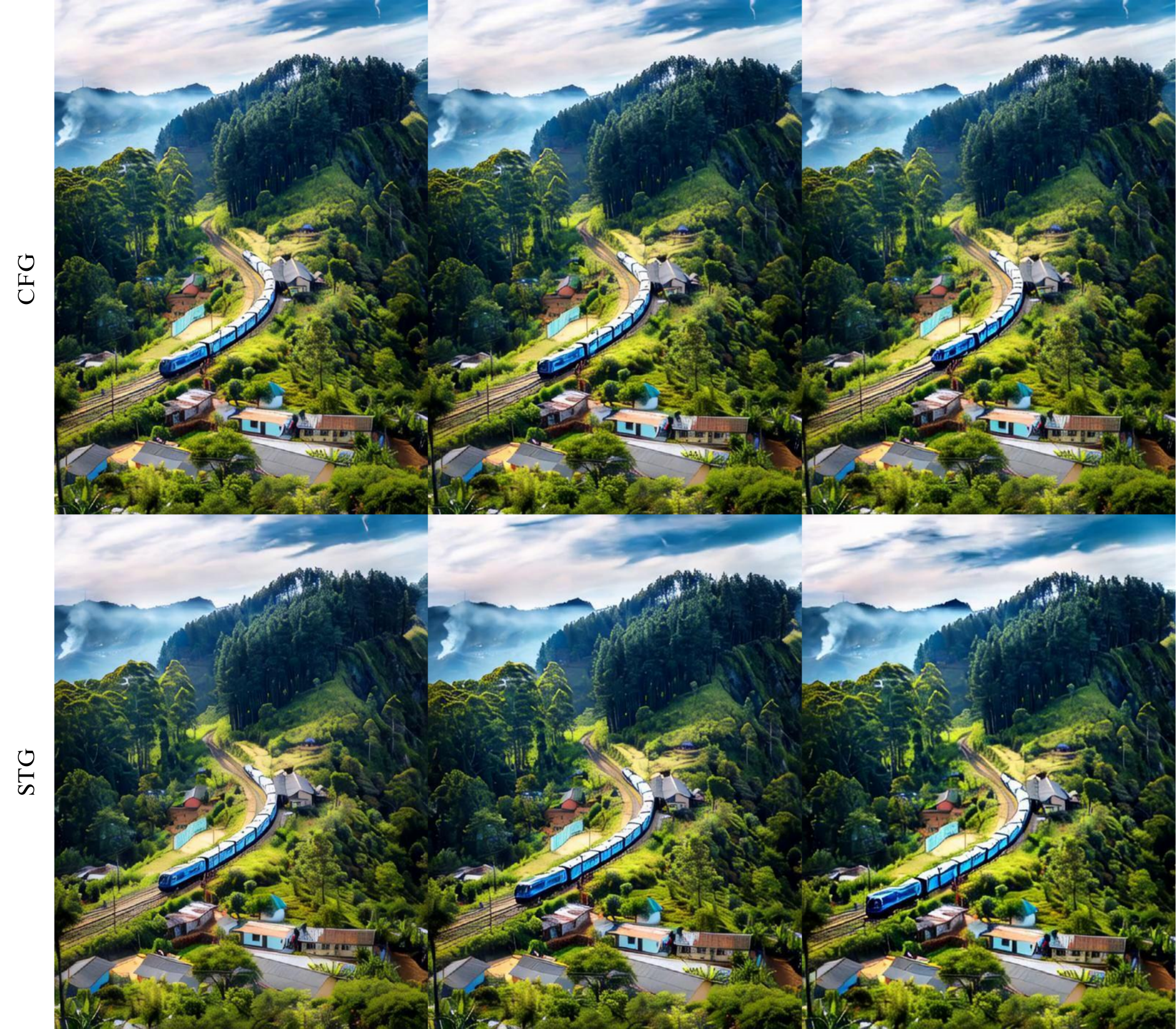}
    \end{subfigure}

    \caption{Qualitative Comparison of Dynamic Degree in SVD~\cite{blattmann2023stable} with and without STG. The results show CFG results in limited object motion, whereas STG mitigates the restrictive effects of CFG, effectively enhancing the motion.}
    \label{fig:dynamic_degree_i2v}
\end{figure}

\end{onecolumn}

\end{document}